*Article*

# Within-Family EfficientNet Scaling under Fixed-Resolution Transfer Learning for Spectral-Domain Optical Coherence Tomography Retinal Disease Classification

**Shkëlqim Sherifi [1] (0009-0006-2227-5533) *, Florim Idrizi [1] (0000-0001-7514-3282), Shpend Ismaili [1]**

[1] University of Tetova; florim.idrizi@unite.edu.mk, shpend.ismaili@unite.edu.mk,
* Correspondence: s.sherifi211185@unite.edu.mk ; Tel.: +389 71697957

## Abstract

**Background/Objectives**: Spectral-domain optical coherence tomography (SD-OCT) provides high-resolution cross-sectional imaging of the retina, while deep convolutional neural networks (CNNs) have demonstrated strong performance for automated retinal disease classification. However, whether the relative performance trajectory of EfficientNet variants transfers from natural-image benchmarks to domain-specific medical imaging remains insufficiently characterized, particularly under a fixed-resolution transfer-learning protocol. This study evaluates the complete EfficientNet-B0 → B7 family under a unified fixed-resolution transfer-learning protocol for four-class SD-OCT retinal disease classification, focusing on the within-family scaling trajectory, class-specific performance, and accuracy-efficiency trade-offs. **Methods**: Eight ImageNet-pretrained EfficientNet variants (B0 → B7), spanning 4.05M-64.10M experimentally instantiated backbone parameters, were evaluated under an identical fixed-resolution protocol. The Kermany four-class SD-OCT dataset was processed using grayscale-to-RGB channel replication, model-embedded EfficientNet preprocessing, stochastic augmentation, and a two-stage progressive fine-tuning strategy comprising frozen-backbone head training followed by partial backbone unfreezing and joint fine-tuning. Performance was evaluated using aggregate classification metrics, class-wise precision, recall and F1 scores, ROC and precision-recall analyses, confusion matrices, training dynamics, inference latency, parameter efficiency, predictive-probability structure, Grad-CAM, and t-SNE representation analysis. **Results**: Test accuracy varied non-monotonically across the EfficientNet family, ranging from 95.93% for B0 to a maximum of 96.95% for B1, with B7 recovering to 96.73% despite its substantially larger parameter count. B1 achieved the highest observed test accuracy (96.95%) and Drusen F1 (0.8780), whereas B7 achieved 96.73% test accuracy, Drusen F1 of 0.8720, and the highest Drusen recall (0.9177). Drusen consistently had the lowest F1 across all eight variants and accounted for 82.2% of the total summed squared deviation in class-specific F1. The F1 ranking is invariant across models, with Normal > CNV > DME > Drusen. The relative ranking of EfficientNet variants also differed between ImageNet and the present OCT evaluation, with B1 ranking first in OCT test accuracy and B7 ranking first in the reported ImageNet hierarchy. The descriptive accuracy-to-parameter ratio, expressed as test accuracy percentage points per million total model parameters, decreased from 19.82 for B0 to 1.48 for B7. The final train-test gap shows a weak, non-significant association with test accuracy ($r = 0.26$, $p = 0.53$). **Conclusion**: Under the evaluated fixed-resolution transfer-learning protocol, increasing EfficientNet scale did not yield monotonic improvement in SD-OCT classification performance. B1 provided the highest observed test accuracy and Drusen F1, whereas B7 exhibited a secondary recovery at substantially greater parameter count and CPU inference latency. Drusen remained the lowest-F1 class across all variants and accounted for 82.2% of the summed squared deviation in class-specific F1. These findings indicate that the downstream response to within-family EfficientNet scaling is task-dependent and protocol-dependent. However, because the downloaded partitions have not been independently audited for cross-partition duplicate images, the reported metrics should be interpreted as conditional on the supplied dataset split.

# 1. Introduction

## *1.1. Clinical and Technical Motivation*

Spectral-domain optical coherence tomography (SD-OCT) provides high-resolution cross-sectional visualization of retinal microstructure and is routinely used to characterize pathological changes associated with choroidal neovascularization (CNV), diabetic macular edema (DME), and Drusen, together with structurally normal retina [1], [2], [3]. Reliable discrimination among these categories requires recognition of structural patterns that may differ in spatial extent, morphology, and visual prominence. These characteristics make SD-OCT a suitable domain for investigating automated image-based retinal disease classification.

Despite advances in automated analysis, expert interpretation remains central to OCT assessment. The increasing volume of retinal imaging, together with inter-reader variability and the limited availability of specialist expertise, motivates computational methods that can provide consistent and scalable decision support [4], [5], [6]. Deep learning-based computer-aided diagnosis (CAD) systems have therefore received substantial attention for automated retinal disease classification, particularly for the four-class CNV/DME/Drusen/Normal benchmark introduced by Kermany et al. [2].

Kermany et al. [2] established this four-class OCT classification task as an important benchmark for subsequent methodological studies and reported that an ImageNet-pretrained InceptionV3 model achieved performance comparable to that of retinal specialists under the conditions of their evaluation. Subsequent work has investigated a wide range of convolutional and hybrid architectures for OCT classification, but direct comparison between reported architectures remains difficult because studies frequently differ in preprocessing, augmentation, optimization, fine-tuning, and evaluation procedures.

## *1.2. Transfer Learning for SD-OCT Classification*

Transfer learning provides a practical framework for adapting ImageNet-pretrained convolutional neural networks to medical-imaging tasks in which labeled target-domain data are more limited than the large-scale datasets used for source-domain pretraining. The effectiveness of this transfer depends, among other factors, on the relationship between the source-domain representations and the visual statistics, spatial structure, and task requirements of the target domain [7], [8], [9], [10].

The distinction between the source and target domains is particularly relevant for SD-OCT. OCT B-scans are grayscale, cross-sectional representations with highly structured retinal anatomy, whereas ImageNet contains natural RGB images characterized by substantially different visual statistics and object-level semantics [1], [2], [11], [12].

Consequently, source-domain performance provides useful information about architectural capability but does not necessarily determine the relative performance of different architectures after adaptation to OCT classification.

This issue becomes particularly important when comparing members of the same architecture family. If larger source-domain models systematically provide stronger representations, one might expect their relative ordering to remain beneficial after transfer. Conversely, if the relationship between architecture scale, representation adaptation, and downstream data is domain-dependent, larger variants may not consistently outperform smaller or intermediate variants. A controlled within-family comparison is therefore useful for separating architectural scale from other sources of methodological variation.

### *1.3. EfficientNet Scaling and the Fixed-Resolution Problem*

EfficientNet was introduced as a systematic approach to neural network scaling based on coordinated changes in network depth, width, and input resolution [7]. The EfficientNet-B0 → B7 family was constructed by progressively increasing a scaling coefficient from the baseline B0 architecture, producing substantially larger variants with progressively greater representational and computational capacity. On ImageNet, the reported Top-1 accuracy increases from 77.1% for B0 to 84.3% for B7 [7].

The original EfficientNet formulation, however, couples architectural scaling with changes in input resolution. The downstream behavior of the family can therefore not be inferred solely from the canonical ImageNet trajectory when the target-domain experiment constrains spatial resolution. In the present study, all variants are evaluated at a fixed input resolution of $224 \times 224$ pixels. The resulting experiment should consequently be understood as a fixed-resolution within-family scaling evaluation, in which the depth and width dimensions vary across B0 → B7 while spatial resolution is held constant.

This constraint is important for two reasons. First, it permits a controlled comparison of progressively larger EfficientNet variants under an identical spatial input representation. Second, it allows the study to examine whether the relative performance ordering associated with EfficientNet scaling on ImageNet is preserved when the models are transferred to a specialized medical-imaging task under a common target-domain protocol.

The central question is therefore not whether EfficientNet scaling improves performance on ImageNet, which has already been established [7], but whether increasing model scale produces a comparable downstream trajectory in SD-OCT classification when resolution, preprocessing, augmentation, training schedule, and evaluation procedures are controlled.

Three outcomes are therefore plausible within the evaluated setting. First, the ImageNet-derived ordering may be preserved after transfer, indicating a monotonic transfer of relative architectural performance. Second, increasing scale may produce diminishing downstream returns, such that larger variants provide progressively smaller improvements. Third, the downstream trajectory may be non-monotonic, with an intermediate variant outperforming one or more substantially larger models. The present study is designed to distinguish among these possibilities empirically.

### *1.4. Research Gap and Research Questions*

Although EfficientNet variants have been applied to ophthalmic and medical-imaging tasks [10], [13], [14], [15], the literature surveyed in this study provides limited evidence on the complete B0 → B7 family evaluated under a common SD-OCT protocol. Existing studies frequently investigate individual EfficientNet variants or restricted subsets of the family, making it difficult to characterize how performance changes across the full evaluated architecture spectrum.

A second unresolved issue concerns the class-specific response to architectural scaling. Aggregate accuracy can remain high while performance differences become concentrated in a particular diagnostic category. This consideration is especially relevant to Drusen, which may present a more challenging precision-recall balance than the other classes in the four-class OCT task. A systematic within-family analysis is therefore required to determine whether scaling improves class-specific performance uniformly or whether the response depends strongly on diagnostic category.

A third methodological issue is comparability. Published OCT studies often differ in preprocessing, augmentation, optimization, fine-tuning, model-selection criteria, and evaluation procedures. Consequently, differences in reported performance cannot always be attributed to architecture alone. The present study addresses this issue through a unified experimental protocol in which the evaluated EfficientNet variants share the same dataset partitions, preprocessing, augmentation, classification head, progressive fine-tuning strategy, and evaluation framework.

Accordingly, the study addresses the following research questions:

1. **How does test accuracy vary across EfficientNet-B0 → B7 under a fixed $224 \times 224$ input resolution, and does the observed trajectory exhibit monotonic, diminishing returns, or non-monotonic behavior?**
2. **How do predictive performance and computational cost jointly vary across EfficientNet-B0 → B7 under the common experimental protocol?**
3. **How does Drusen classification respond to within-family scaling across B0 → B7, and how are changes in Drusen precision and recall reflected in the resulting F1 trajectory?**

*1.5. Contributions*

This study makes four principal contributions:

1. **Controlled within-family evaluation.** We provide a systematic evaluation of EfficientNet-B0 → B7 for four-class SD-OCT retinal disease classification under a common fixed-resolution transfer-learning and progressive fine-tuning protocol.
2. **Characterization of the downstream B0 → B7 scaling trajectory**. We systematically characterize the observed performance trajectory across the complete evaluated EfficientNet family and examine whether the source-domain ImageNet ordering is preserved after transfer to SD-OCT.
3. **Class-specific scaling analysis.** We decompose model performance at the class level and characterize the behavior of Drusen, including its F1 trajectory and associated precision-recall changes across the evaluated architecture spectrum.
4. **Accuracy-efficiency characterization.** We jointly evaluate predictive performance and computational cost using experimentally instantiated parameter counts and measured CPU inference latency to characterize the observed accuracy-efficiency trade-off across the family.

The analysis is conducted under a matched experimental configuration using the same nominal random seed across all eight variants. This common seed is used as a control for the specified stochastic configuration rather than as a substitute for repeated-seed assessment of variability.

The remainder of this paper is organized as follows. **Section 2** reviews deep learning for ophthalmic and OCT analysis, CNN-based retinal classification, EfficientNet scaling and transfer learning, attention and hybrid CNN-Transformer architectures, and the resulting research gaps. **Section 3** describes the dataset, preprocessing, and augmentation pipeline, EfficientNet-B0 → B7 configurations, classification head, progressive fine-tuning protocol, reproducibility controls, computational environment, and evaluation methodology. **Section 4** presents the empirical results, while **Section 5** interprets the observed scaling behavior, class-specific effects, computational trade-offs, and qualitative representation analyses. **Section 6** summarizes the principal findings and identifies directions for future investigation.

## 2. Literature Review

### *2.1. Deep Learning for Ophthalmic and OCT Image Analysis*

Deep learning has become an established methodological framework for ophthalmic image analysis, with convolutional neural networks (CNNs) providing the dominant architectural paradigm in many early and intermediate studies [1], [16]. Reviews of ophthalmic deep learning have documented the progression from early convolutional architectures toward residual, densely connected, Inception-based, lightweight, ensemble, and more recently Transformer-based models, while identifying limited labeled data, class imbalance, domain heterogeneity, and interpretability as persistent methodological challenges [1], [16], [17].

Within OCT analysis, Kermany et al. [2] established the four-class CNV/DME/Drusen/Normal classification task that subsequently became an important benchmark for automated retinal disease classification. Their work demonstrated the potential of ImageNet-pretrained deep networks for OCT classification and provided a methodological reference point for subsequent architectural studies. Later investigations have expanded the range of models and training strategies applied to OCT, reflecting a broader transition from proof-of-concept classification toward comparative, interpretable, and deployment-oriented evaluation.

More recent research has increasingly examined domain-specific representation learning as an alternative or complement to ImageNet transfer. Du et al. [18] introduced RETFound, a retinal foundation model pretrained using self-supervised learning on a large collection of retinal images, illustrating the growing interest in learning representations directly from ophthalmic data. This development is relevant to the present study because it reinforces the broader question of how effectively ImageNet-derived representations and architectural scaling transfer to retinal imaging tasks. At the same time, work addressing clinical translation has emphasized the importance of external validation, inter-device heterogeneity, uncertainty estimation, and prospective evaluation when moving beyond retrospective benchmark studies [4].

### *2.2. CNN Architectures for Retinal OCT Classification*

The post-Kermany literature encompasses a diverse range of CNN architectures and computational strategies. This architectural diversity provides evidence that OCT classification performance depends not only on the task and dataset but also on the architectural representation and associated training protocol.

Residual architectures have been widely investigated for retinal OCT classification. Li et al. [19] compared VGG-16, InceptionV3, and ResNet-50 for retinal disease detection, while Asif et al. [20] evaluated residual learning for OCT disease diagnosis. Pang et al. [21] subsequently investigated residual architectures for macular-degeneration classification using a dedicated preprocessing strategy. Collectively, these studies demonstrate the applicability of residual representations to OCT while also illustrating that reported performance is coupled to the preprocessing and optimization procedures used by each study.

Dense connectivity represents a second architectural direction. Akinniyi et al. [22] incorporated DenseNet-style components within a multi-stage ensemble for retinal OCT classification, while Sotoudeh-Paima et al. [23] investigated multi-scale convolutional representations for AMD classification. These studies emphasize feature reuse and multi-scale representation as mechanisms for extracting diagnostically relevant information from retinal images.

A complementary line of research has focused on lightweight and deployment-oriented CNNs. Alqudah et al. [11] proposed AOCT-NET as a grayscale-oriented architecture for SD-OCT classification, while OctNET and other lightweight CNN approaches [24], [25], [26] emphasized reduced model complexity and computational requirements. This literature demonstrates that predictive accuracy is not the only criterion relevant to OCT model selection and motivates explicit consideration of computational cost alongside classification performance. Sherifi et al. [27] introduced ED126-A, a compact four-class CNN trained from scratch on the Kermany OCT dataset. With 1.24 million parameters, it achieved 95.03% accuracy and a macro-F1 of 0.9205, while identifying Drusen as the main source of classification errors. This work provides a from-scratch reference baseline complementary to the present evaluation of ImageNet-pretrained EfficientNet variants.

Ensemble and multi-stage approaches provide another methodological direction. Ho et al. [28] reported improved retinal-image classification using ensembles of CNN backbones, while Peng et al. [12] incorporated denoising and multi-scale residual processing for OCT classification. Comparative studies such as those of Mohan et al. [29] and Singh et al. [30] further illustrate the value of evaluating multiple architectures under comparative conditions. However, these studies generally focus on differences between architectural families rather than on how the performance of a single architecture family changes progressively with model scale. This distinction motivates the within-family comparison undertaken in the present study.

### *2.3. EfficientNet, Transfer Learning, and Architecture Scaling*

Tan and Le [7] introduced EfficientNet as a systematic framework for scaling convolutional neural networks by jointly adjusting network depth, width, and input resolution. Rather than independently increasing a single architectural dimension, the EfficientNet formulation coordinates these dimensions through a compound scaling strategy derived from the EfficientNet-B0 baseline. The resulting B0 $\rightarrow$ B7 family provides a structured sequence of increasingly large models with progressively greater representational and computational capacity.

On ImageNet, the reported Top-1 accuracy increases from 77.1% for EfficientNet-B0 to 84.3% for EfficientNet-B7 [7]. This source-domain performance trajectory provides a natural reference for investigating whether the relative ordering of the family is preserved after transfer to a specialized target domain. The present study is particularly concerned with this transfer behavior because the downstream experiment fixes input resolution rather than reproducing the original depth-width-resolution scaling configuration.

EfficientNet employs MBConv blocks combining inverted bottleneck transformations, depthwise convolution, squeeze-and-excitation (SE) channel recalibration, and nonlinear activation functions [7], [31], [32]. SE provides channel-wise recalibration based on global activation statistics, and is natively incorporated into the EfficientNet architecture [7], [32]. Although such channel-wise recalibration may be useful when discriminative information is distributed unevenly across feature channels, its effect is inherently task and architecture dependent and should not be interpreted as a demonstrated mechanism for any particular OCT disease class without direct experimental evidence [33].

EfficientNet variants have subsequently been investigated in ophthalmic and medical-imaging applications. Wang et al. [10] evaluated EfficientNet-B0-B5 for multi-label fundus-image classification, providing evidence that compound-scaled representations can transfer from natural-image pretraining to retinal photography. Abbas et al. [15]

proposed an optimized EfficientNet-based architecture for hypertensive and diabetic retinopathy classification, while Arora et al. [14] evaluated an EfficientNet-B3/B5 ensemble for diabetic retinopathy grading. For OCT specifically, D. and Grace [13] investigated EfficientNet-B3 with neuromorphic similarity-score enhancement. These studies establish the applicability of EfficientNet to retinal imaging, but they do not directly characterize the complete B0 → B7 family for four-class SD-OCT classification under a common experimental protocol.

The literature on transfer learning further indicates that source-task performance does not necessarily determine downstream performance uniformly across target domains. The degree of representation transfer depends on the relationship between source and target visual statistics, the amount of target-domain data, and the extent of adaptation induced by fine-tuning [34], [35], [36], [37]. Accordingly, increasing source-domain model capacity cannot be assumed to produce a monotonic downstream benefit after transfer to OCT.

Architecture-level scaling should also be distinguished from parameter-reduction strategies such as pruning. Poyatos et al. [34] demonstrated that evolutionary pruning can reduce redundant channels while maintaining competitive transfer-learning performance. Pruning and EfficientNet scaling therefore address model efficiency from different directions: pruning modifies a given architecture to reduce redundancy, whereas within-family scaling changes architectural capacity. The present study focuses on the latter and examines how predictive performance and computational cost change across the full evaluated B0 → B7 spectrum.

### *2.4. Attention, Hybrid CNN-Transformer, and Emerging OCT Architectures*

Attention mechanisms provide an important complementary direction in medical image analysis. Squeeze-and-excitation networks introduced channel-wise feature recalibration [32], which is already integrated into the EfficientNet MBConv design [7]. The Convolutional Block Attention Module (CBAM) subsequently extended attention to both channel and spatial dimensions by sequentially modeling feature importance and spatial localization [38]. In retinal imaging, spatially aware attention is potentially relevant when discriminative anatomical structures are localized. Farag et al. [39] reported improvements from CBAM-augmented DenseNet-121 for diabetic retinopathy grading, while Huang et al. [40] investigated attention mechanisms specifically for retinal OCT classification. These studies provide context for evaluating Efficient Net's built-in channel recalibration, but they do not establish whether additional spatial attention would alter the within-family scaling behavior.

Recent OCT research has also expanded toward hybrid CNN-Transformer and Transformer-based architectures. Ma et al. [41] proposed HCTNet, combining convolutional feature extraction with Transformer-based global-context modeling for retinal OCT classification. A broader architectural perspective is provided by Sherifi [42], who reviewed the evolution of deep learning architectures from multilayer Perceptrons and conventional convolutional neural networks (CNNs) to Vision Transformers and hybrid models. This architectural progression highlights the complementary roles of convolutional feature extraction and attention-based global-context modeling, providing a broader conceptual context for the development and evaluation of CNN-based and hybrid approaches in medical image analysis. Dutta et al. [43] developed Conv-ViT, integrating convolutional and Vision Transformer representations for retinal disease detection. Broader medical-imaging reviews have emphasized dataset size, pretraining, and task characteristics as important determinants of Transformer performance [44], [45]. Takahashi et al. [9] further reported that CNNs can remain competitive with Vision

Transformers on moderate-sized medical-imaging datasets, supporting the continued relevance of convolutional inductive biases when labeled data are limited.

The broader research landscape is also moving toward volumetric OCT, explainable and neuromyotonic approaches, standardized datasets, and external validation. Hammou et al. [46] investigated 2D/3D retinal OCT classification, while Miladinovic et al. [47] proposed a neuromyotonic framework for explainable OCT analysis. Kulyabin et al. [48] introduced the OCTDL dataset to support reproducible benchmarking, and Leingang et al. [49] evaluated automated AMD detection using real-world multi-center OCT data. These developments illustrate a transition from isolated benchmark accuracy toward broader consideration of representation quality, interpretability, reproducibility, and generalizability.

*2.5. Research Gap Synthesis*

The reviewed literature establishes three unresolved issues that motivate the present study.

**First**, although EfficientNet variants have been applied to ophthalmic and medical-imaging problems, including fundus classification and individual retinal disease tasks [10], [13], [14], [15], the literature reviewed here provides limited evidence on the complete EfficientNet-B0 → B7 family evaluated for four-class SD-OCT classification under a common preprocessing, optimization, fine-tuning, and evaluation protocol. Existing studies therefore provide evidence for EfficientNet applicability but do not adequately characterize the downstream performance trajectory across the full evaluated family.

**Second**, the class-specific response to architectural scaling remains insufficiently characterized. Existing OCT studies commonly emphasize aggregate classification performance, whereas the consequences of increasing model scale for individual diagnostic categories have received less systematic attention [12], [22], [29]. A within-family evaluation at the class level is therefore required to determine whether scaling produces broadly uniform improvements or whether performance variation is concentrated classes.

**Third**, direct comparison of previously reported OCT architectures is complicated by differences in preprocessing, augmentation, optimization, fine-tuning, model selection, and evaluation procedures. Such methodological differences make it difficult to attribute performance differences specifically to architecture. A controlled within-family benchmark that holds these factors constant therefore provides a more direct way to characterize the relationship between architectural scale and downstream OCT classification performance.

These research gaps are also consistent with the findings of a recent scoping review by Sherifi [50], which examined 55 research records on CNN-based OCT and retinal-imaging classification published between 2017 and 2026. The review identified persistent methodological limitations, including the insufficient availability of controlled first-principles baselines, incomplete characterization of ImageNet-to-OCT transfer effectiveness, limited reporting of class-sensitive evaluation metrics, and insufficient investigation of learned representation geometry and cross-scanner generalization. These observations further emphasize the importance of controlled architectural comparisons, class-specific performance analysis, and representation-level evaluation in retinal OCT classification.

Together, these gaps motivate the present evaluation of EfficientNet-B0 → B7 under a fixed $224 \times 224$ resolution and a common transfer-learning protocol. The study consequently examines the observed downstream scaling trajectory, its class-specific consequences, and the associated accuracy-efficiency trade-offs while explicitly distinguishing the constrained fixed-resolution experiment from the canonical three-dimensional EfficientNet compound-scaling formulation.

# 3. Methodology and Methods

## *3.1. Dataset and Data Partition*

The four-class spectral-domain optical coherence tomography (SD-OCT) dataset was introduced by Kermany et al. [2] and obtained through the Labeled Optical Coherence Tomography (OCT) release distributed on Kaggle by Anirudh CV [51]. The downloaded release contained 109,309 images across four classes: choroidal neovascularization (CNV), diabetic macular edema (DME), Drusen, and Normal. The training, validation, and test partitions supplied with the Kaggle release were retained unchanged across all eight EfficientNet variants, comprising 76,515, 21,861, and 10,933 images, respectively. The class-specific counts for the training, validation, and test subsets were 26,218/7,491/3,746 for CNV, 8,118/2,319/1,161 for DME, 6,206/1,773/887 for Drusen, and 35,973/10,278/5,139 for Normal [51]. No additional random partition was generated. The nominal random seed of 42 described in Section 3.6 applied to the specified stochastic training components, not to dataset allocation. The original partition-generation procedure and patient-level separation were not independently verified in this study. The supplied dataset partitions have the same image counts as the OCT2018 variant examined by Marciniak and Stankiewicz [52], who reported 2,171 validation images and 1,112 test images duplicated in the training partition. However, the files used in the present experiments have not been independently compared with the files examined in that audit, and the published overlap counts should therefore not be interpreted as direct measurements of the files used in this study. Although images in the designated test directory were not intentionally supplied to the training pipeline or used for model selection, the presence or absence of cross-partition duplicate images in the downloaded files remains unverified. Consequently, independence of test-image content from training data cannot currently be established. The dataset is substantially imbalanced, with Normal as the most prevalent class and Drusen as the least prevalent class. To compensate for this imbalance during optimization, class-weighted categorical cross-entropy was used for both training stages.

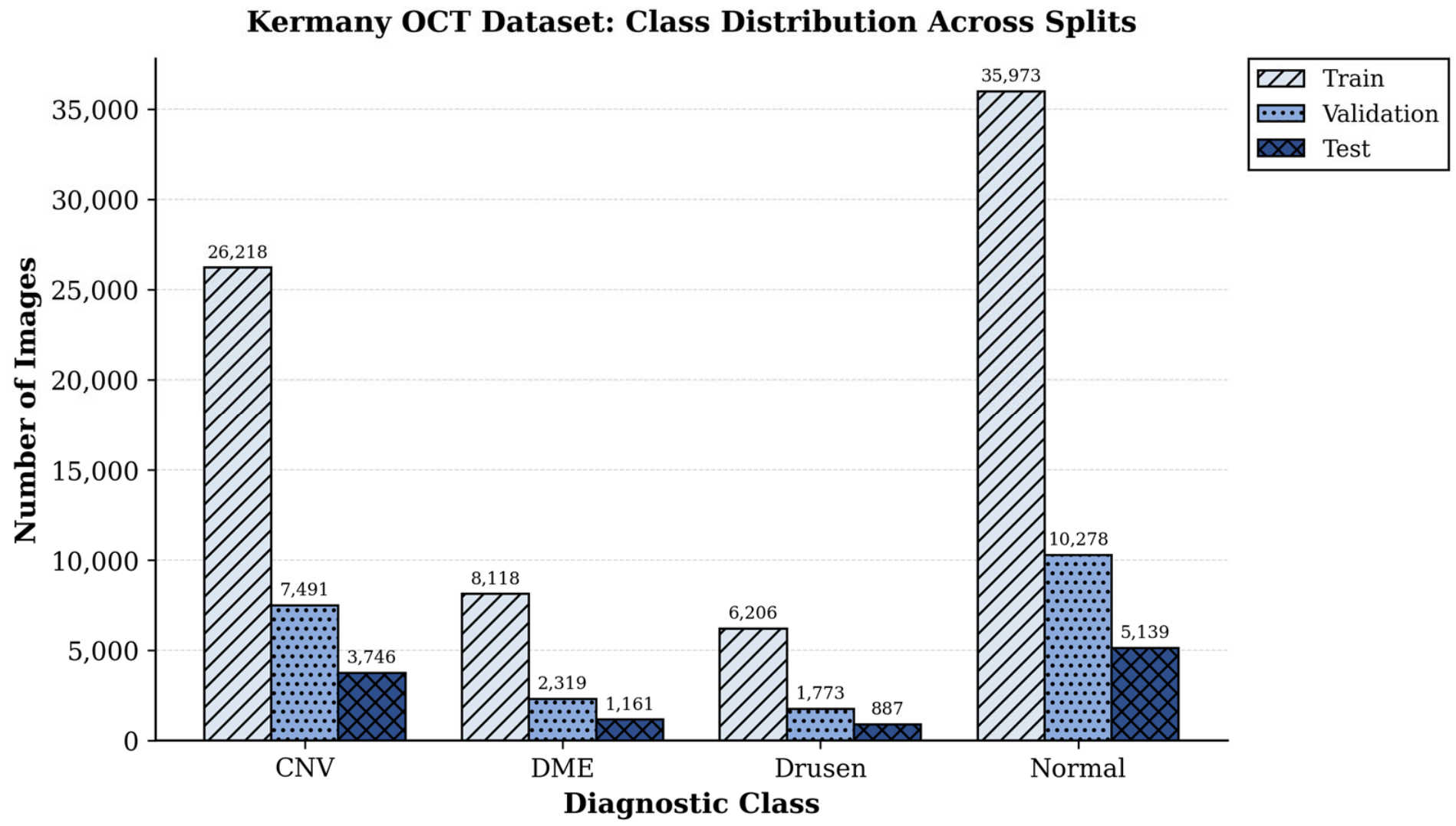

**Figure 1.** Kermany SD-OCT dataset.

All images were resized to $224 \times 224$ pixels using bilinear interpolation. The same resizing procedure was applied uniformly across all EfficientNet variants. This fixed-resolution constraint was applied uniformly to B0 $\rightarrow$ B7 so that the experimental comparison evaluated differences associated with EfficientNet architectural scaling while holding input resolution constant. Thus, the present experiment should be interpreted as a fixed-resolution within-family scaling study rather than as a reproduction of the complete depth-width-resolution compound-scaling trajectory defined in the original EfficientNet formulation.

### *3.2. Unified Preprocessing and Augmentation*

All eight EfficientNet variants were subjected to the same input preprocessing, augmentation policy, optimization configuration, training schedule, and nominal random seed. This common protocol was used to minimize methodological differences between variants and to provide a controlled basis for the within-family comparison.

#### 3.2.1. Input Pipeline and Preprocessing

All OCT images were processed as grayscale single-channel images and standardized to a spatial resolution of $224 \times 224$ pixels. The resulting tensor has the form: $X \in R^{B\times 224\times 224\times 1}$, where (B) denotes the batch dimension. Because all evaluated ImageNet-pretrained backbones require three-channel input, the single grayscale channel was deterministically replicated along the channel dimension: $X_3 = [X, X, X] \in R^{B\times 224\times 224\times 3}$.

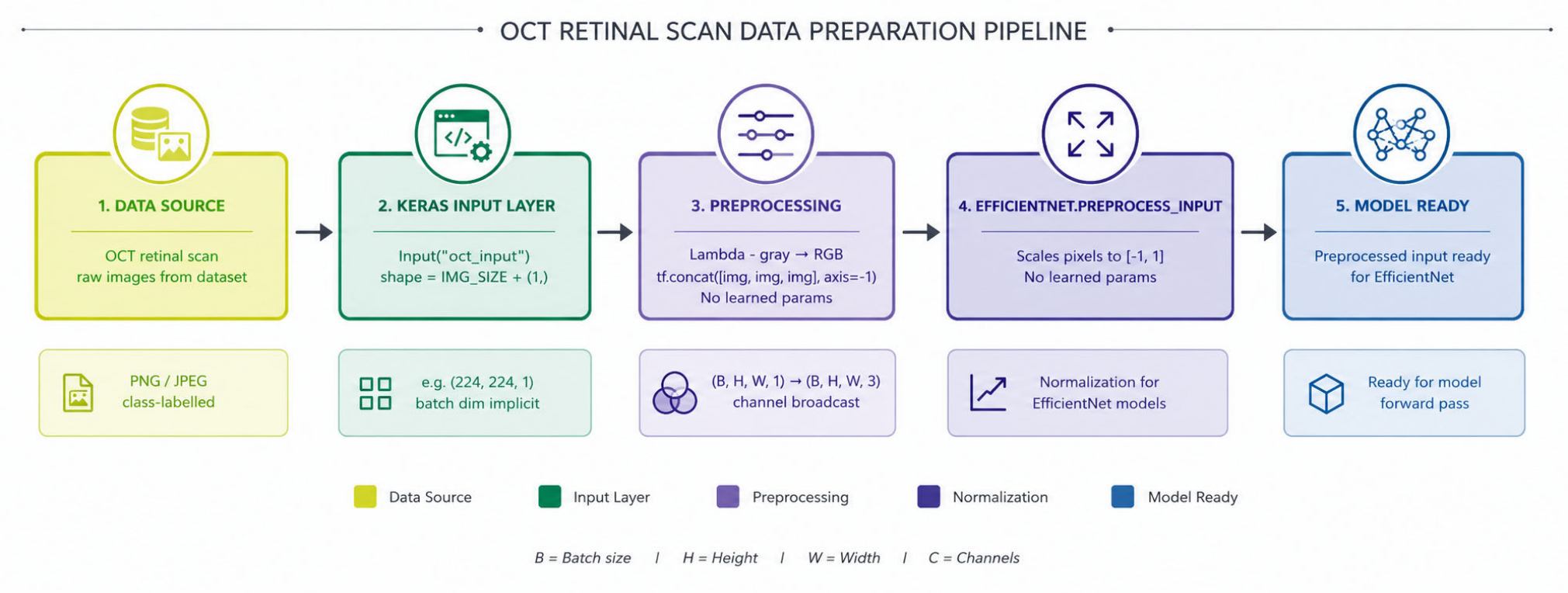

**Figure 2.** EfficientNet pipeline input.

The channel-replication operation is parameter-free and does not introduce additional learnable weights. It allows the grayscale OCT images to be presented to the pretrained ImageNet backbones using the expected three-channel input representation.

After channel replication, the resulting three-channel images were provided to the TensorFlow/Keras EfficientNet models in the expected input range, with preprocessing handled by the model implementation.

#### 3.2.2. Data Augmentation

Augmentation was applied online during training and disabled during validation and test inference. The policy consisted of the following transformations:

1. **Horizontal random flip**: horizontal mirroring with the nominal seed set to 42. Vertical flipping was not applied because retinal-layer orientation is diagnostically meaningful.
2. **Random rotation**: rotation within $\pm\mathbf{5}^{\circ}$, using nearest-neighbor border filling.
3. **Random translation**: translation within $\pm\mathbf{5}\%$ of the image dimensions, using nearest-neighbor border filling.
4. **Random zoom**: zoom perturbation within $\pm 5\%$, again using nearest-neighbor border filling.
5. **Random contrast**: contrast variation with a factor of 0.15 and nominal seed 42.

The augmentation policy was intentionally restricted to small geometric and photometric perturbations to preserve the principal anatomical structure of the OCT B-scans while introducing limited variability during optimization.

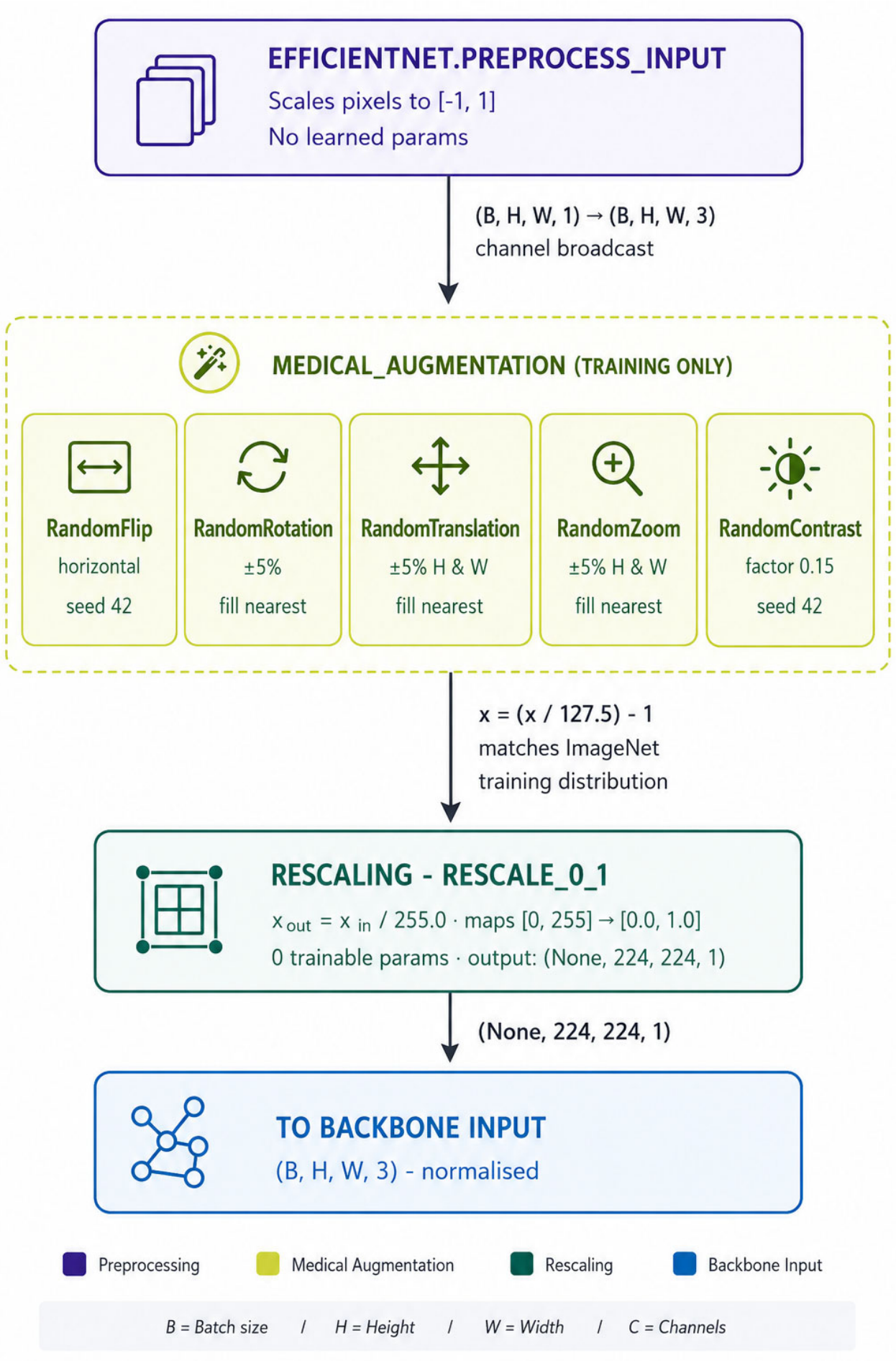

**Figure 3.** EfficientNet pipeline Augmentation.

### *3.3. EfficientNet-B0 → B7 Architectures*

The eight evaluated models belong to the EfficientNet family introduced by Tan and Le [7]. The family is organized around a common convolutional design incorporating a stem convolution, MBConv stages, a top projection convolution, and the classification head described in Section 3.4. MBConv blocks employ inverted-bottleneck expansion, depthwise convolution, squeeze-and-excitation (SE) channel recalibration, and pointwise projection. The backbone uses the Swish activation function, while the SE mechanism performs channel-wise recalibration through global pooling, a bottleneck transformation, and sigmoid gating.

The canonical EfficientNet family defines model scale through coordinated depth, width, and input-resolution changes indexed by the scaling coefficient (ϕ):

$$[d(\phi) = \alpha^{\phi}, \qquad w(\phi) = \beta^{\phi}, \qquad r(\phi) = \gamma^{\phi}], \tag{1}$$

where (α), (β), and (γ) are the scaling coefficients specified by the EfficientNet formulation. The canonical B0 → B7 configurations therefore differ in depth, width, and resolution. In the present study, however, the resolution term was deliberately constrained to 224 × 224 for every variant. Consequently, the experimental comparison evaluates the within-family changes in architectural depth and width at a common spatial input resolution and should not be described as a full reproduction of the original three-dimensional compound-scaling law.

Table 1 summarizes the canonical EfficientNet architectural specifications used as the reference family definition.

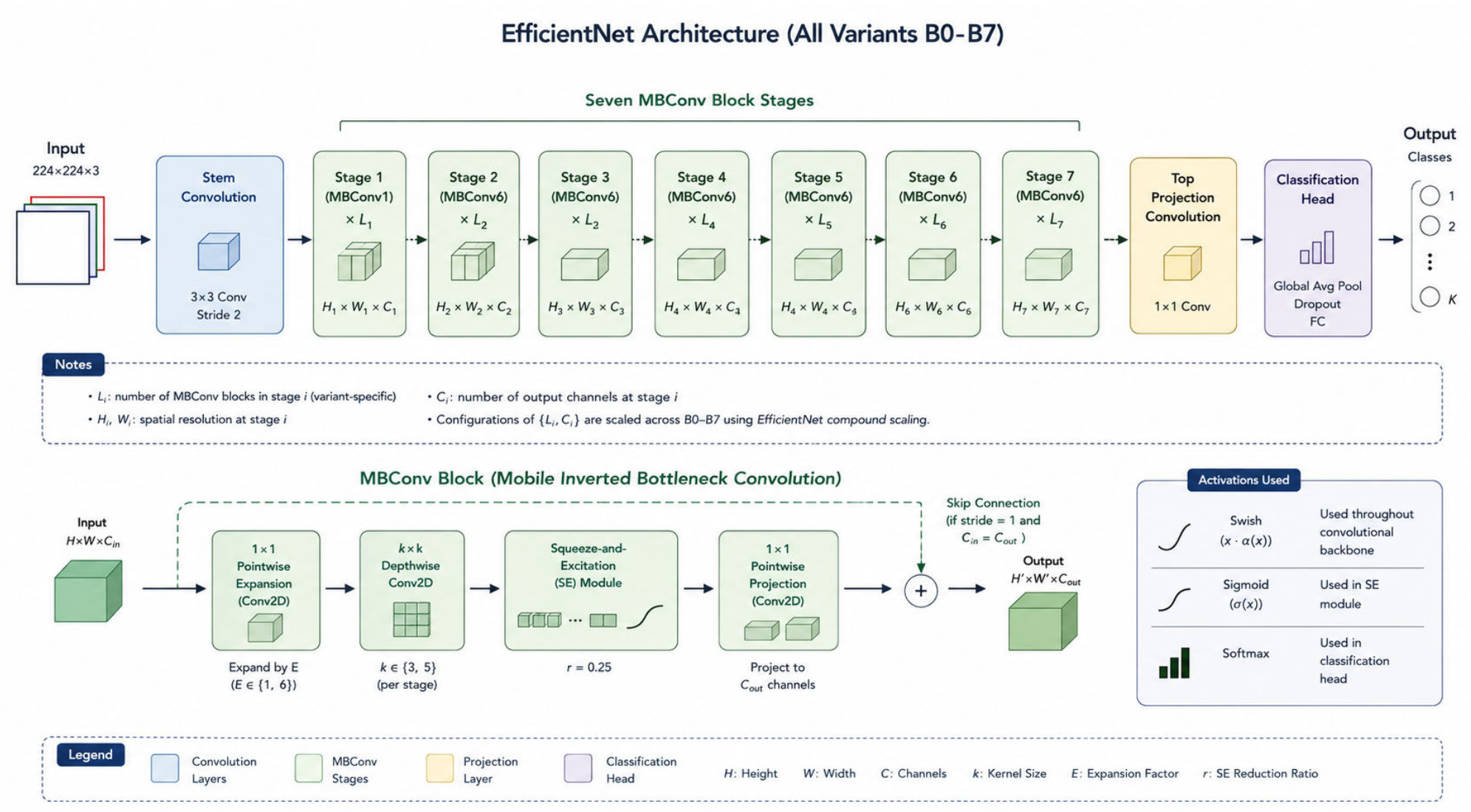

**Figure 4.** EfficientNet Architecture Shared Design Principles.

**Table 1.** Canonical EfficientNet Architectural Specifications

| Variant | φ | Width(w) | Depth(d) | Canonical Model Params | Top Conv Filters | ImageNet Top 1 | MBConv Blocks | 5 × 5 Kernel Stages |
|---|---|---|---|---|---|---|---|---|
| B0 | 0 | 1 | 1 | 5.3M | 1,280 | 77.1% | 16 | 3, 5, 6 |
| B1 | 1 | 1 | 1.1 | 7.8M | 1,280 | 79.1% | 23 | 3, 5, 6 |
| B2 | 2 | 1.1 | 1.2 | 9.2M | 1,408 | 80.1% | 23 | 3, 5, 6 |
| B3 | 3 | 1.2 | 1.4 | 12M | 1,536 | 81.6% | 26 | 3, 5, 6 |
| B4 | 4 | 1.4 | 1.8 | 19M | 1,792 | 82.9% | 32 | 3, 5, 6 |
| B5 | 5 | 1.6 | 2.2 | 30M | 2,048 | 83.6% | 39 | 3, 5, 6 |
| B6 | 6 | 1.8 | 2.6 | 43M | 2,304 | 84.0% | 45 | 3, 5, 6 |
| B7 | 7 | 2.0 | 3.1 | 66M | 2,560 | 84.3% | 55 | 3, 5, 6 |

* **Note**: These canonical parameter counts include the original ImageNet classification head. The experimentally instantiated backbone parameter counts reported elsewhere in this paper correspond to the Keras EfficientNet backbones initialized with include_top=False, from which the original ImageNet classification head was removed.

The canonical parameter counts in Table 1 are reference architectural specifications and should be distinguished from the experimentally instantiated parameter counts obtained for the present implementation. The experimentally instantiated backbone parameter counts used throughout the Results section were 4.05M, 6.58M, 7.77M, 10.78M, 17.67M, 28.51M, 40.96M, and 64.10M for B0 → B7, respectively. The corresponding total model parameter counts, including the task-specific classification head, were 4.84M, 7.37M, 8.63M, 11.71M, 18.73M, 29.70M, 42.28M, and 65.54M. These experimentally instantiated values, rather than the rounded canonical counts, are used when reporting model size, parameter-normalized quantities, and computational cost in the Results section.

All variants retained the EfficientNet MBConv design, including the $5 \times 5$ depthwise-convolution stages in Blocks 3, 5, and 6. The architectural scaling comparison therefore preserved the underlying EfficientNet family design while varying the prescribed family-scale configuration under the fixed $224 \times 224$ input constraint.

### *3.4. Custom Classification Head*

The ImageNet classification head was removed from each backbone using *include_top=False*, and the same task-specific classification-head topology and hyperparameters were attached to all eight variants. Because the dimensionality of the backbone feature output varies across EfficientNet variants, the instantiated parameter counts of the first fully connected layer, and hence of the complete classification head, varies accordingly. The head consisted of:

1. GlobalAveragePooling2D to collapse the spatial feature representation into a one-dimensional feature vector.
2. a fully connected layer with 512 units and ReLU activation, followed by Batch Normalization and Dropout with (p=0.5).
3. a fully connected layer with 256 units and ReLU activation, followed by Batch Normalization and Dropout with (p=0.3) and
4. a four-unit softmax output layer producing the class probabilities for CNV, DME, Drusen, and Normal.

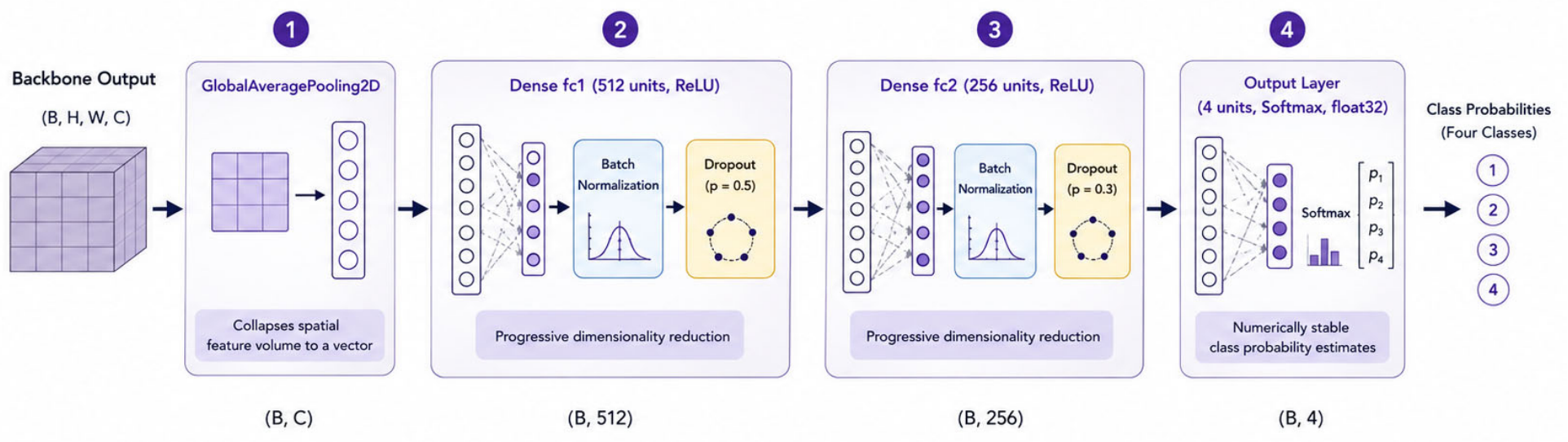

**Figure 5.** Custom Four-Class Classification Head (Shared Across All Evaluated Backbones).

### *3.5. Progressive Transfer-Learning and Fine-Tuning Protocol*

All eight variants were trained using the same two-stage progressive transfer-learning procedure.

#### 3.5.1. Stage 1: Frozen-Backbone Head Training

The backbones were instantiated using the TensorFlow/Keras EfficientNet-B0-B7 implementations in TensorFlow 2.20.0, with weights=*'imagenet'* and include_top=*False*. The corresponding preprocessing behavior of the selected Keras implementation was retained. All backbone parameters were frozen, while the task-specific classification head was trained for a maximum of 15 epochs. The backbone was invoked with training=*False*, maintaining the *Batch Normalization* layers in inference mode during frozen-backbone training.

Optimization used AdamW with learning rate $\eta_1 = 1 \times 10^{-3}$ and weight decay $\lambda_1 = 1 \times 10^{-4}$.

The training objective was categorical cross-entropy with label smoothing $\alpha = 0.1$. EarlyStopping and ModelCheckpoint both monitored validation loss ($val_loss$). EarlyStopping used a patience of 5 epochs, while ModelCheckpoint retained the checkpoint corresponding to the lowest observed validation loss. ReduceLROnPlateau monitored validation loss (*val_loss*), using a factor of 0.5, a patience of 2 epochs, and a minimum learning rate of $10^{-6}$. CSVLogger was used to record the training history.

#### 3.5.2. Transition to Partial Backbone Fine-Tuning

After Stage 1, the checkpoint corresponding to the lowest observed validation loss was restored. The backbone was then divided into approximately equal lower and upper portions. The lower 50%, comprising the stem and early MBConv blocks, remained frozen, whereas the upper 50%, comprising later MBConv stages and the top convolution, was unfrozen for joint fine-tuning. The optimizer state was reinitialized at this transition.

#### 3.5.3. Stage 2: Joint Fine-Tuning

During Stage 2, trainable Batch Normalization layers in the unfrozen upper backbone and task-specific classification head were updated during training, whereas Batch Normalization layers in the frozen lower backbone remained non-trainable. For final evaluation, the checkpoint corresponding to the lowest observed Stage 2 validation loss was restored before inference on the held-out test set. AdamW was used with a reduced learning rate $\eta_2 = 1 \times 10^{-5}$ and weight decay $\lambda_2 = 1 \times 10^{-5}$.

The loss remained categorical cross-entropy with label smoothing $\alpha = 0.1$.

EarlyStopping and ModelCheckpoint both monitored validation loss ($val_loss$). EarlyStopping used a patience of 6 epochs, while ModelCheckpoint retained the checkpoint corresponding to the lowest observed validation loss. ReduceLROnPlateau monitored validation loss (*val_loss*), using a factor of 0.5, a patience of 3 epochs, and a minimum learning rate of $10^{-7}$. CSVLogger was retained to record the training history.

The class weights were calculated from the training-set class frequencies using:

$$W_c = \frac{N_{total}}{N_{classes} N_c}, \tag{2}$$

where $N_{\text{total}}$ is the number of training images, $N_{\text{classes}} = 4$, and $N_c$ is the number of training images in class $c$. This produced approximate weights of 0.73 for CNV, 2.36 for DME, 3.08 for Drusen, and 0.53 for Normal. The same weights were applied in both training stages.

### *3.6 Reproducibility and Random-Seed Control*

A common nominal random seed of 42 was used for all eight EfficientNet variants. The same seed and experimental configuration were applied across B0 → B7 for the reported stochastic components, including model-head initialization and stochastic data augmentation. The dataset partitions, preprocessing pipeline, optimization settings, batch size, training schedule, and classification head were also kept constant across variants.

The common seed provides a matched stochastic condition for the within-family comparison, but it should not be interpreted as an estimate of stochastic uncertainty or between-run variability. Use of the same nominal seed does not establish that different architectures experienced identical random-number realizations in an architectural sense, nor does it establish that the relative ordering of closely performing variants would remain unchanged under independent seeds. Accordingly, the comparison is descriptive with respect to the single evaluated seed.

Because only one seed was evaluated, repeated independent training runs would be required to quantify run-to-run variability and assess the statistical robustness of small differences between variants, particularly where the observed performance differences are small.

**Table 2.** Training Configuration Summary

| Parameter | Stage 1 | Stage 2 |
|---|---|---|
| Max epochs | 15 | 30 |
| Optimizer | AdamW | AdamW |
| Learning rate ($\eta$) | $1 \times 10^{-3}$ | $1 \times 10^{-5}$ |
| Weight decay (λ) | $1 \times 10^{-4}$ | $1 \times 10^{-5}$ |
| Loss function | CCE + Label Smoothing | CCE + Label Smoothing |
| Batch size | 16 | 16 |
| Mixed precision | float16 compute -> float32 output | float16 compute -> float32 output |
| Class weights | Balanced (inverse frequency) | Balanced (inverse frequency) |
| Backbone state | Fully frozen | Top 50% unfrozen, bottom 50% frozen |
| Random seed | Fixed and identical across B0 → B7 | Fixed and identical across B0 → B7 |
| EarlyStopping patience | 5 | 6 |
| ReduceLROnPlateau factor | 0.5 | 0.5 |
| ReduceLROnPlateau patience | 2 | 3 |
| Min learning rate | $1 \times 10^{-6}$ | $1 \times 10^{-7}$ |
| ModelCheckpoint monitor | Validation loss | Validation loss |
| EarlyStopping monitor | Validation loss | Validation loss |

### *3.7 Computational Environment*

**Table 3.** Hardware Infrastructure

| **Resource** | **Specification** | **Role** |
|---|---|---|
| Training GPU | NVIDIA Tesla T4 (16 GB VRAM) | Training was performed on an NVIDIA Tesla T4 GPU with 16 GB VRAM in the reported Google Colab environment. |
| Inference CPU | Intel(R) Core(TM) i7-6700HQ CPU @ 2.60 GHz (2.59 GHz) - CPU inference-latency measurement | CPU inference latency measurement |
| Framework | TensorFlow 2.20.0 with Keras API | Model definition, training, export |
| Mixed Precision | float16 compute / float32 output | Memory optimization |
| Batch Size | 16 (all variants) | Unified across |
| Input Resolution | 224 × 224 (all variants) | Isolates scaling from resolution φ |

Training was performed in the reported Google Colab environment using an NVIDIA Tesla T4 GPU with 16 GB of VRAM. The software environment used TensorFlow 2.20.0 with the Keras API, while mixed-precision training employed float16 computation with float32 output. The common training batch size was 16 and the input resolution was fixed at 224 × 224 for all variants.

**Table 4.** Compute Requirements Across φ

| Variant | Backbone Params | Approx. Training Time (Stage 1+2) | Peak VRAM |
|---|---|---|---|
| B0 | 4.05M | ~340 min* | ~6 GB* |
| B1 | 6.58M | ~365 min* | ~7 GB* |
| B2 | 7.77M | ~375 min* | ~8 GB* |
| B3 | 10.78M | ~375 min* | ~10 GB* |
| B4 | 17.67M | ~400 min* | ~12 GB* |
| B5 | 28.51M | ~500 min* | ~15 GB* |
| B6 | 40.96M | ~675 min* | ~16 GB (near T4 limit)* |
| B7 | 64.10M | ~845 min* | ~16 GB (near T4 memory limit; mixed-precision training used)* |

* **Approximate values**; per-epoch timing was not logged due to multi-stage training across sessions, which overwrote prior logs. VRAM reflects observed, not fixed, usage.

Observed training time and peak VRAM utilization were recorded as approximate operational measurements for each variant. Across B0 → B7, approximate Stage 1 + Stage 2 training times ranged from approximately 340 min to 845 min, while observed peak VRAM usage increased from approximately 6 GB for B0 to approximately 16 GB for B6 and B7. Because per-epoch timing was not retained consistently across the multi-session training process, these values should be interpreted as approximate experiment-level measurements rather than standardized throughput benchmarks.

**Table 5.** CPU Inference Latency and Single-Image Throughput for EfficientNet B0 → B7

| Variant | Backbone Params | Total Params | Mean Latency (ms) | Median Latency (ms) | Throughput (img/s) |
|---|---|---|---|---|---|
| B0 | 4.05M | 4.84M | 407.48 | 345.51 | 2.454 |
| B1 | 6.58M | 7.37M | 512.79 | 474.43 | 1.950 |
| B2 | 7.77M | 8.63M | 479.46 | 474.48 | 2.086 |
| B3 | 10.78M | 11.71M | 544.27 | 543.42 | 1.837 |
| B4 | 17.67M | 18.73M | 683.46 | 681.20 | 1.463 |
| B5 | 28.51M | 29.70M | 844.98 | 841.37 | 1.183 |
| B6 | 40.96M | 42.28M | 971.32 | 970.17 | 1.030 |
| B7 | 64.10M | 65.54M | 1247.88 | 1232.05 | 0.801 |

* **Approximate values;** Measurements were taken on a non-dedicated machine and may be affected by concurrent background processes and non-exclusive use of available hardware resources.

Inference latency was measured independently on an Intel Core i7-6700HQ CPU at 2.60 GHz under Windows 10. Batch size was set to 1. Each model underwent 50 warm-up runs followed by 200 timed inference runs using float32 inference at the standardized input size. Mean and median latency were recorded for each variant. Single-image throughput was calculated as:

$$\mathrm{T} = \frac{1000}{\mathrm{L}}, \quad (3)$$

where $(L)$ is the mean latency in milliseconds and $(T)$ is expressed in images per second.

The latency measurements were obtained on a non-dedicated CPU system and may therefore be affected by concurrent background processes and non-exclusive use of hardware resources. Consequently, the reported latency values characterize the computational cost under the specified experimental environment and should not be interpreted as hardware-independent deployment times.

### *3.8. Evaluation Metrics and Statistical Analysis*

Evaluation was structured around six complementary analytical axes: aggregate classification performance, class-specific performance, confusion-matrix analysis, predictive-probability structure, spatial attribution, and embedding geometry. This framework was designed to characterize both overall classification performance and class-dependent behavior across the EfficientNet scaling spectrum.

**Table 6.** Six-Axis Evaluation Framework

| Axis | Metrics | Diagnostic Purpose |
|---|---|---|
| 1. Aggregate Scalar | Accuracy, Weighted F1, Macro F1, Balanced Accuracy, MCC, Cohen's Kappa | Holistic performance ranking across φ |
| 2. Per-Class Decomposition | Precision, Recall, F1, per class | Identifying class-specific scaling effects (esp. Drusen) |
| 3. Confusion Matrix | Absolute counts, row-normalized recall | Quantifying inter-class confusion patterns |
| 4. Top-k Probability Structure | C(y_true) vs. C(competitor); max(C) vs. second-max(C) | Qualitative characterization of predictive-probability organization across φ |
| 5. Spatial Attribution | Grad-CAM heatmaps, multi-layer Grad-CAM, feature maps | Anatomical grounding of model decisions |
| 6. Embedding Geometry | t-SNE (perplexity 30, 1000 iterations) | Qualitative visualization of learned representation geometry across $\varphi$ |

#### 3.8.1. Aggregate Classification Performance

Aggregate performance was assessed using test accuracy, balanced accuracy, Matthew's correlation coefficient (MCC), Cohen's kappa, weighted F1, and macro F1. Training accuracy and the final train-test accuracy gap were additionally reported to provide a descriptive view of generalization behavior.

The final train-test gap was defined as the difference between training accuracy and test accuracy for the corresponding variant. This quantity was interpreted descriptively and was not treated as a direct measure of a specific overfitting mechanism. Pearson correlation was used to quantify the association between final train-test gap and test accuracy across the eight evaluated variants. Because the analysis contains only eight architectural observations, the resulting correlation should be interpreted as descriptive rather than as evidence of a general population-level relationship.

#### 3.8.2. Class-Specific Performance

For each class, precision, recall, and F1 score were calculated separately for CNV, DME, Drusen, and Normal. Class-specific analyses were used to determine whether changes in architectural scale affected all diagnostic categories similarly or whether performance variation was concentrated classes.

For Drusen, additional analysis decomposed performance into precision, recall, sensitivity, specificity, and the precision-recall gap,

$$\Delta_{PR} = |\text{Recall} - \text{Precision}|, \quad (4)$$

using the corresponding class-level values reported for each model. The association between the Drusen precision-recall gap and Drusen F1 was summarized using Pearson correlation across the eight EfficientNet variants. This analysis was treated as a descriptive cross-variant association rather than as evidence of causal dependence.

#### 3.8.3. Within-Family F1 Variability Decomposition

To quantify how class-specific F1 variability was distributed across the four diagnostic categories, the within-family squared deviation for each class ($c$) was calculated as:

$$S_c = \sum_{i=1}^{8}\left(F1_{c,i} - \overline{F1}_c\right)^2, \quad (5)$$

where ($F1_{c,i}$) is the F1 score of class ($c$) for EfficientNet variant ($i$), and ($\overline{F1}_c$) is the mean F1 score of that class across B0 → B7. The percentage contribution of class ($c$) to the total summed squared deviation was then defined as:

$$V_c = 100\frac{S_c}{\sum_k S_k}, \quad (6)$$

This calculation was derived from the class-specific F1 values in Table 9 and was used to quantify the concentration of observed F1 variation among the four diagnostic classes.

To quantify within-model disparity between the highest- and lowest-performing diagnostic classes, the best-to-worst class-F1 difference was defined as:

$$\Delta_i = \max_c\left(F1_{c,i}\right) - \min_c\left(F1_{c,i}\right), \quad (7)$$

where $F1_{c,i}$ denotes the F1 score for class $c$ under EfficientNet variant $i$.

3.8.4. Source-Domain versus Target-Domain Ranking

The relative ordering of EfficientNet variants on ImageNet and on the OCT, task was compared using Spearman rank correlation between the ImageNet Top-1 ranking and the OCT test-accuracy ranking. The analysis was used to characterize whether the source-domain architectural hierarchy was preserved after transfer to the target medical-imaging task.

Because only eight variants were evaluated, the resulting rank correlation was interpreted cautiously and as an exploratory assessment of ranking consistency rather than as a definitive test of transferability.

3.8.5. Accuracy-to-Parameter Ratio

A descriptive parameter-normalized quantity was calculated as:

$$E_{\text{param}} = \frac{\text{Test}_{\text{accuracy}}(\%)}{\text{Total}_{\text{parameters}}\ (\text{M})}, \quad (8)$$

This quantity expresses test accuracy in percentage points per million total parameters. It is used only as a descriptive efficiency indicator for the evaluated variants and is not treated as a standardized efficiency metric or as evidence of a general scaling law.

3.8.6. ROC and Precision-Recall Analysis

One-vs-rest receiver operating characteristic (ROC) and precision-recall (PR) curves were evaluated separately for the four diagnostic classes. ROC-AUC and PR-AUC were examined to distinguish threshold-independent discrimination from class-specific precision-recall behavior. Particular attention was given to Drusen because its PR-AUC varied more substantially across variants than its ROC-AUC, indicating that discrimination and precision-recall behavior should not be treated as interchangeable quantities.

3.8.7. Confusion-Matrix Analysis

Confusion matrices were generated for each EfficientNet variant using the fixed held-out test set. Both absolute class counts, and row-normalized recall patterns were examined to characterize inter-class error structure. This analysis was used to identify whether differences among models reflected broad degradation across the four classes or were concentrated in specific error patterns, particularly those involving Drusen.

3.8.8. Predictive-Probability Structure

Predictive probabilities were examined using the confidence-triangle representation described in the Results section. For each test observation, the probability assigned to the true class and the highest competing class probability were considered when characterizing the organization of model confidence across the eight variants.

This analysis is qualitative and is intended to describe predictive-probability structure rather than to provide a formal assessment of probabilistic calibration. No claim of improved or degraded calibration is therefore made without a dedicated calibration metric.

3.8.9. Spatial Attribution and Representation Geometry

Grad-CAM and related feature-map analyses were used to provide qualitative spatial attribution of representative predictions. These analyses were intended to examine whether highlighted image regions were anatomically plausible and whether representative correct and incorrect predictions displayed different spatial patterns.

t-SNE was used to visualize the geometry of learned feature embeddings, using perplexity 30 and 1000 iterations. The resulting embeddings were interpreted qualitatively to examine representation organization across classes and model variants. Because t-SNE is a nonlinear visualization method, the plots were not treated as quantitative evidence of class separability or monotonic improvement in representation quality.

## 4. Results

All eight EfficientNet variants (B0 → B7) were evaluated under the identical experimental protocol described in Sections 3. Results are presented sequentially in terms of aggregate performance, class-wise performance, ROC and precision-recall characteristics, confusion-matrix structure, training dynamics, computational efficiency, predictive-probability structure, spatial attribution, and representation geometry. The analyses therefore characterize the scaling response at both aggregate and class-specific levels, followed by complementary error, efficiency, and representation analyses. The evaluated variants contained 4.05M-64.10M backbone parameters, and all models were trained using the same preprocessing, augmentation, classification head, and progressive fine-tuning protocol.

### 4.1. Aggregate Performance and Scaling Trajectory

Table 7 reports test accuracy, balanced accuracy, Matthew's correlation coefficient (MCC), Cohen's kappa, weighted F1, macro F1, training accuracy, and the final train-test gap for all eight EfficientNet variants. Aggregate test performance varied non-monotonically across B0 → B7. Test accuracy increased by 1.02 percentage points from B0 to B1, declined at B2, partially recovered at B3, declined again at B4, increased at B5, decreased slightly at B6, and recovered to 96.73% at B7, remaining 0.22 percentage points below B1.

**Table 7.** Full-Spectrum Aggregate Performance

| $\varphi$ | Model | Back-bone Param | Total param | Test Acc. | Bal Acc. | MCC | Cohen K | Wt F1 | Macro F1 | Train Acc. | Train-Test Gap |
|---|---|---|---|---|---|---|---|---|---|---|---|
| 0 | B0 | 4.05M | 4.84M | 0.9593 | 0.9448 | 0.9374 | 0.9372 | 0.9601 | 0.9313 | 0.9646 | 0.53pp |
| 1 | B1 | 6.58M | 7.37M | 0.9695 | 0.9554 | 0.9528 | 0.9527 | 0.9698 | 0.9476 | 0.9808 | 1.13pp |
| 2 | B2 | 7.77M | 8.63M | 0.9624 | 0.9508 | 0.9423 | 0.9421 | 0.9630 | 0.9369 | 0.9707 | 0.83pp |
| 3 | B3 | 10.78M | 11.71M | 0.9652 | 0.9518 | 0.9463 | 0.9461 | 0.9657 | 0.9409 | 0.9801 | 1.49pp |
| 4 | B4 | 17.67M | 18.73M | 0.9601 | 0.9506 | 0.9389 | 0.9386 | 0.9611 | 0.9334 | 0.9781 | 1.80pp |
| 5 | B5 | 28.51M | 29.70M | 0.9643 | 0.9530 | 0.9451 | 0.9450 | 0.9649 | 0.9398 | 0.9782 | 1.39pp |
| 6 | B6 | 40.96M | 42.28M | 0.9636 | 0.9538 | 0.9441 | 0.9439 | 0.9644 | 0.9388 | 0.9816 | 1.80pp |
| 7 | B7 | 64.10M | 65.54M | 0.9673 | 0.9578 | 0.9497 | 0.9495 | 0.9678 | 0.9445 | 0.9847 | 1.74pp |

Figure 6 presents the complete aggregate performance, while Figure 7 plots test accuracy against model size. The test-accuracy sequence from B0 to B7 is 95.93%, 96.95%, 96.24%, 96.52%, 96.01%, 96.43%, 96.36%, and 96.73%, respectively. B1 therefore achieved the highest test accuracy at 96.95%, representing a 1.02-percentage-point improvement over B0. B2 subsequently decreased to 96.24%, B3 partially recovered to 96.52%, and B4 decreased again to 96.01%. The subsequent variants showed a modest rebound, with B5 and B6 reaching 96.43% and 96.36%, respectively, before B7 recovered to 96.73%.

The observed test-accuracy trajectory can be summarized in four descriptive intervals. From B0 to B1, test accuracy increased by 1.02 percentage points, representing the largest positive change between adjacent variants. Across B2-B4, test accuracy decreased at B2, partially recovered at B3, and decreased again to 96.01% at B4, the lowest value among B1-B7. B5 increased to 96.43%, followed by a small decrease to 96.36% at B6; both remained below B1. B7 recovered to 96.73%, remaining 0.22 percentage points below B1 despite a substantially larger parameter count.

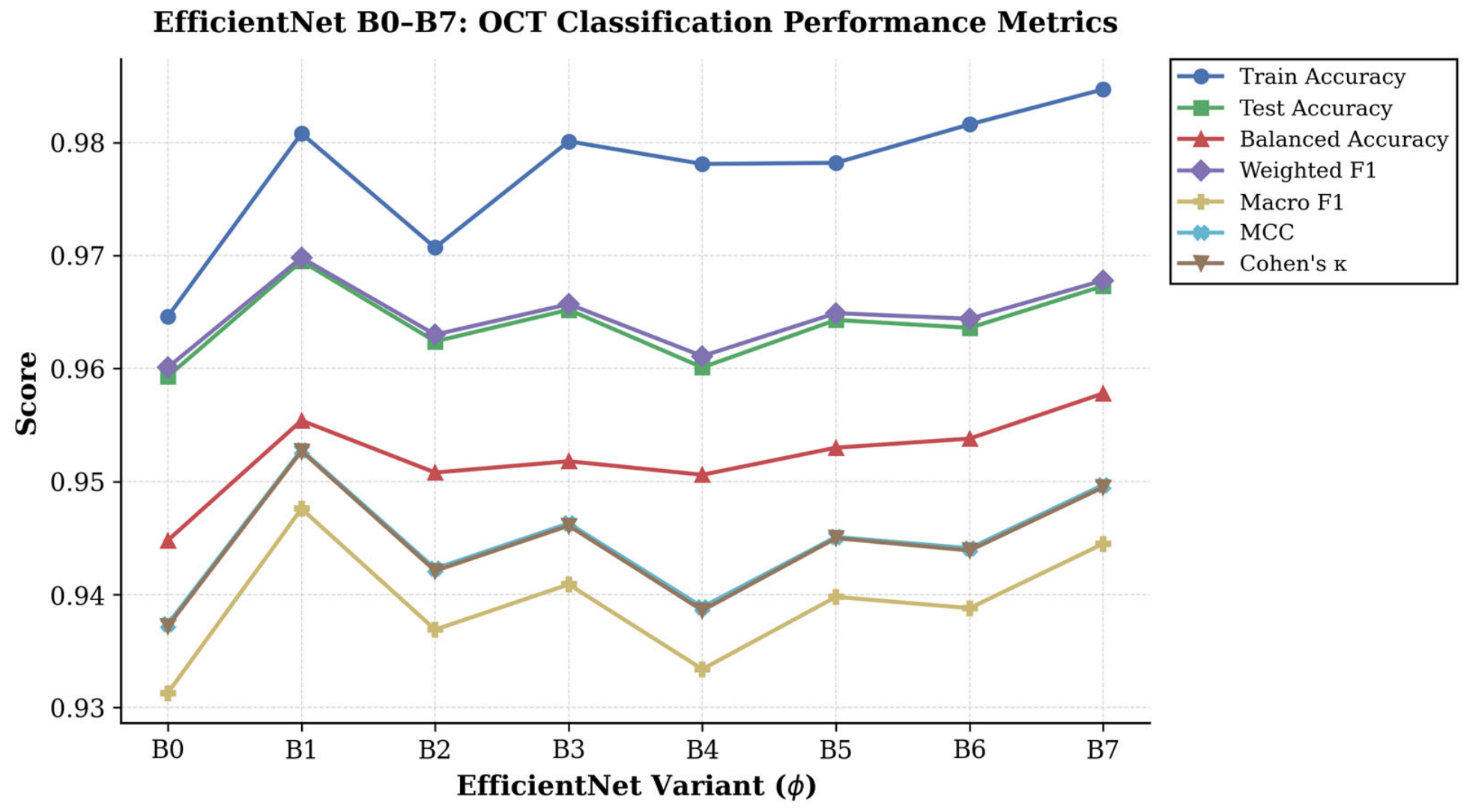

**Figure 6.** Full-Spectrum Aggregate Performance.

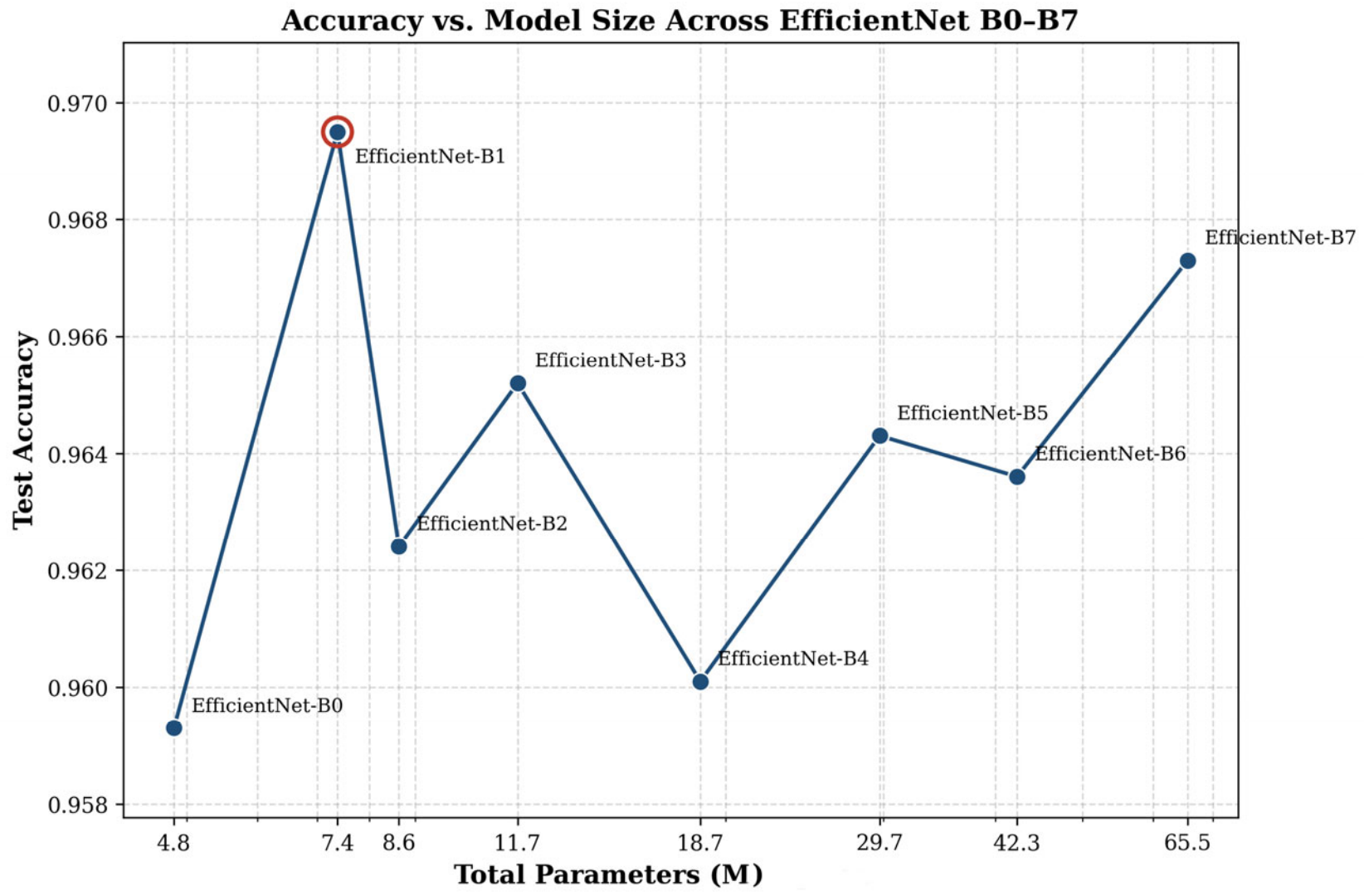

**Figure 7.** Accuracy vs. Model Size Across EfficientNet B0 → B7.

The test-accuracy ranking observed on OCT differed from the ImageNet ranking reported in Table 8. On ImageNet, accuracy increases systematically from B0 through B7, with B7 ranked first. In the present OCT experiment, B1 ranked first in test accuracy, followed by B7, B3, B5, B6, B2, B4, and B0.

**Table 8.** ImageNet and OCT Test-Accuracy Rankings Across EfficientNet-B0 → B7

| Rank | ImageNet Top-1 | ImageNet Top-5 | OCT (this study) Test |
|---|---|---|---|
| 1 | B7 (84.3%) | B7 (97.0%) | B1 (96.95%) |
| 2 | B6 (84.0%) | B6 (96.8%) | B7 (96.73%) |
| 3 | B5 (83.6%) | B5 (96.7%) | B3 (96.52%) |
| 4 | B4 (82.9%) | B4 (96.4%) | B5 (96.43%) |
| 5 | B3 (81.6%) | B3 (95.7%) | B6 (96.36%) |
| 6 | B2 (80.1%) | B2 (94.9%) | B2 (96.24%) |
| 7 | B1 (79.1%) | B1 (94.4%) | B4 (96.01%) |
| 8 | B0 (77.1%) | B0 (93.3%) | B0 (95.93%) |

These results show that the relative ranking of EfficientNet variants differed between ImageNet and the present OCT evaluation. B7 ranked first on the ImageNet accuracy measures reported in Table 8, whereas B1 achieved the highest OCT test accuracy. B1 therefore achieved the highest test-accuracy value with substantially fewer parameters than B7.

The final train-test gaps ranged from 0.53 percentage points for B0 to 1.80 percentage points for B4 and B6. B1 exhibited a 1.13-percentage-point gap, while B2 showed the second-smallest gap at 0.83 percentage points. The Pearson correlation between the final train-test gap and test accuracy was weak and not statistically significant ($r=0.26$, $p=0.53$). Thus, these data do not provide evidence of an association between the two quantities within the eight evaluated variants.

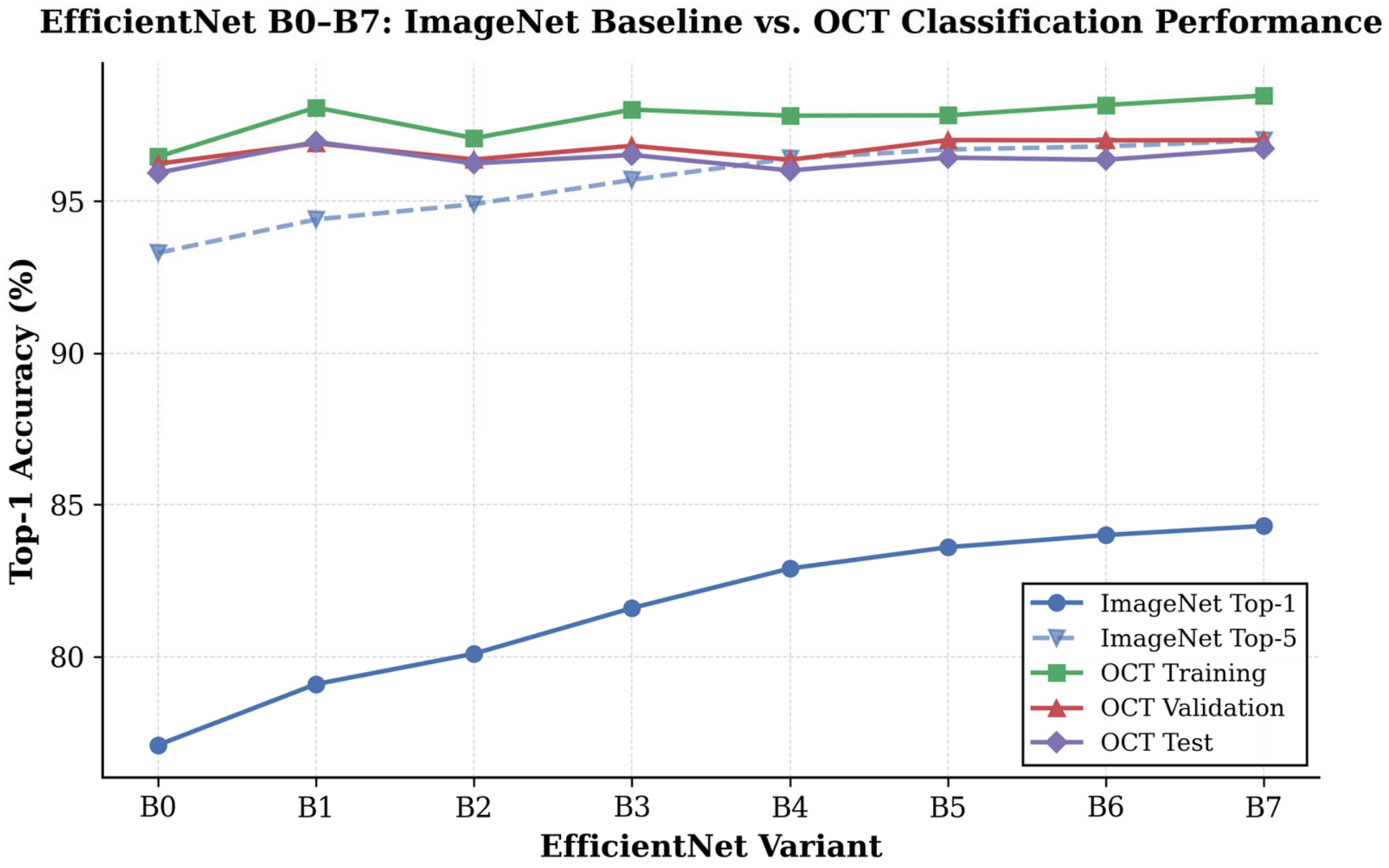

**Figure 8.** ImageNet vs. OCT Scaling Rank Comparison Accuracy.

### *4.2. Per-Class Performance*

Table 9 decomposes precision, recall, and F1 across CNV, DME, Drusen, and Normal for all eight EfficientNet variants. The class-wise metrics show that the response to evaluated within-family scaling trajectory varied across diagnostic categories. Drusen had the lowest F1 value for every evaluated variant, whereas CNV and Normal exhibited smaller F1 ranges. DME showed intermediate variation across the scaling spectrum.

**Table 9.** Full-Spectrum Per-Class F1, Precision, and Recall Trajectory: B0 → B7

| Model | CNV F1 | CNV Prec. | CNV Rec. | DME F1 | DME Prec. | DME Rec. | Drusen F1 | Drusen Prec. | Drusen Rec. | Normal F1 | Normal Prec. | Normal Rec. |
|---|---|---|---|---|---|---|---|---|---|---|---|---|
| B0 | 0.9669 | 0.9821 | 0.9522 | 0.9401 | 0.9207 | 0.9604 | 0.8373 | 0.7900 | 0.8906 | 0.9807 | 0.9855 | 0.9761 |
| B1 | 0.9783 | 0.9875 | 0.9693 | 0.9502 | 0.9387 | 0.9621 | 0.8780 | 0.8493 | 0.9087 | 0.9838 | 0.9859 | 0.9817 |
| B2 | 0.9729 | 0.9813 | 0.9648 | 0.9408 | 0.9112 | 0.9724 | 0.8544 | 0.8162 | 0.8963 | 0.9796 | 0.9895 | 0.9698 |
| B3 | 0.9724 | 0.9847 | 0.9605 | 0.9510 | 0.9417 | 0.9604 | 0.8571 | 0.8129 | 0.9064 | 0.9829 | 0.9861 | 0.9798 |
| B4 | 0.9709 | 0.9900 | 0.9525 | 0.9411 | 0.9195 | 0.9638 | 0.8429 | 0.7826 | 0.9132 | 0.9789 | 0.9848 | 0.9730 |
| B5 | 0.9732 | 0.9866 | 0.9602 | 0.9412 | 0.9126 | 0.9716 | 0.8628 | 0.8251 | 0.9042 | 0.9818 | 0.9876 | 0.9761 |
| B6 | 0.9759 | 0.9861 | 0.9658 | 0.9515 | 0.9395 | 0.9638 | 0.8489 | 0.7914 | 0.9154 | 0.9790 | 0.9879 | 0.9702 |
| B7 | 0.9739 | 0.9858 | 0.9624 | 0.9476 | 0.9232 | 0.9733 | 0.8720 | 0.8306 | 0.9177 | 0.9844 | 0.9909 | 0.9780 |

For CNV, F1 varied only from 0.9669 to 0.9783, with B1 obtaining the highest value. Normal F1 similarly remained within a narrow range of 0.9789-0.9844, with B7 obtaining the highest value. DME showed somewhat greater variability, ranging from 0.9401 for B0 to 0.9515 for B6. By contrast, Drusen F1 exhibited the largest variation across the family.

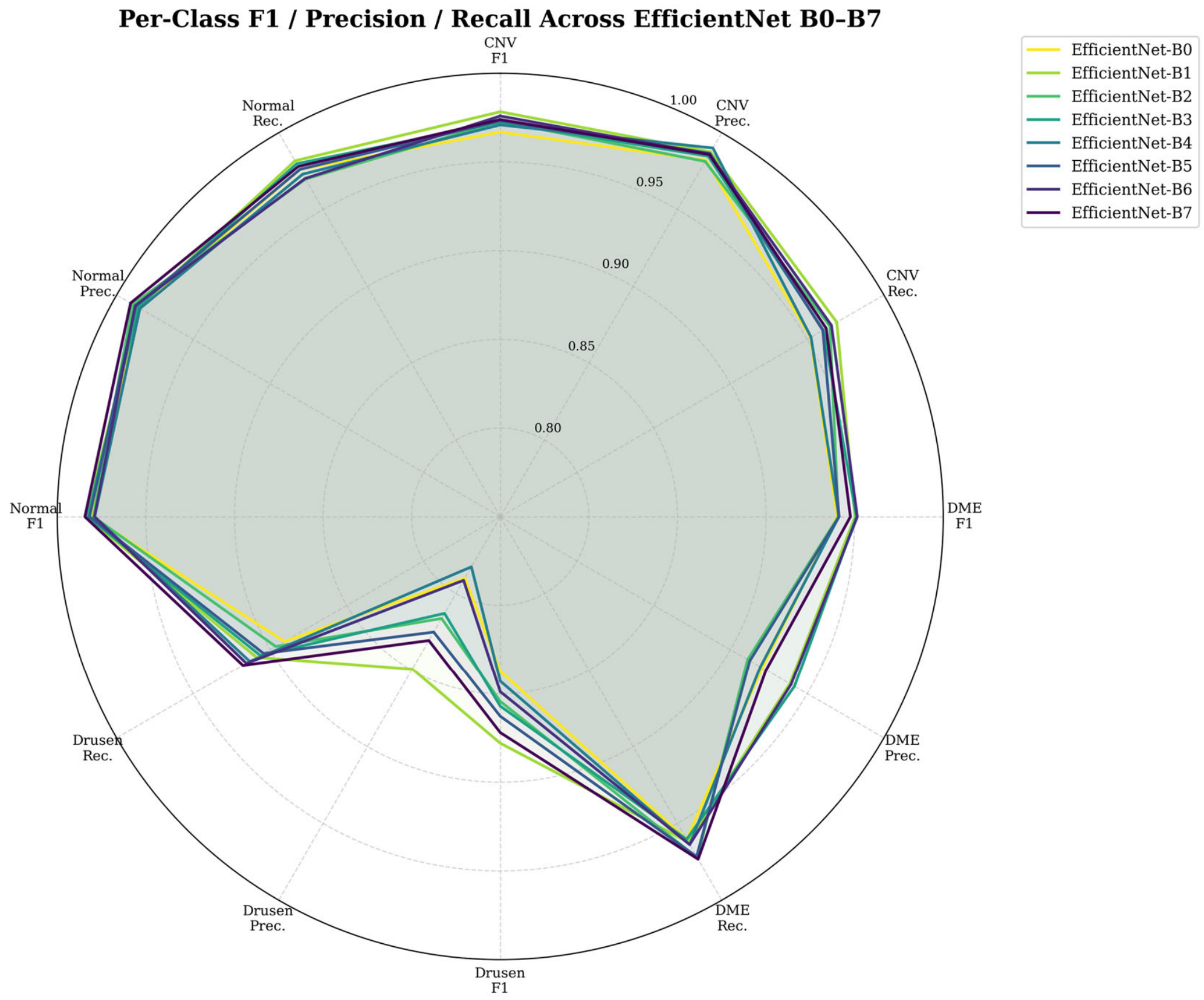

**Figure 9.** Full-Spectrum Per-Class F1, Precision, and Recall Trajectory Across EfficientNet-B0 → B7.

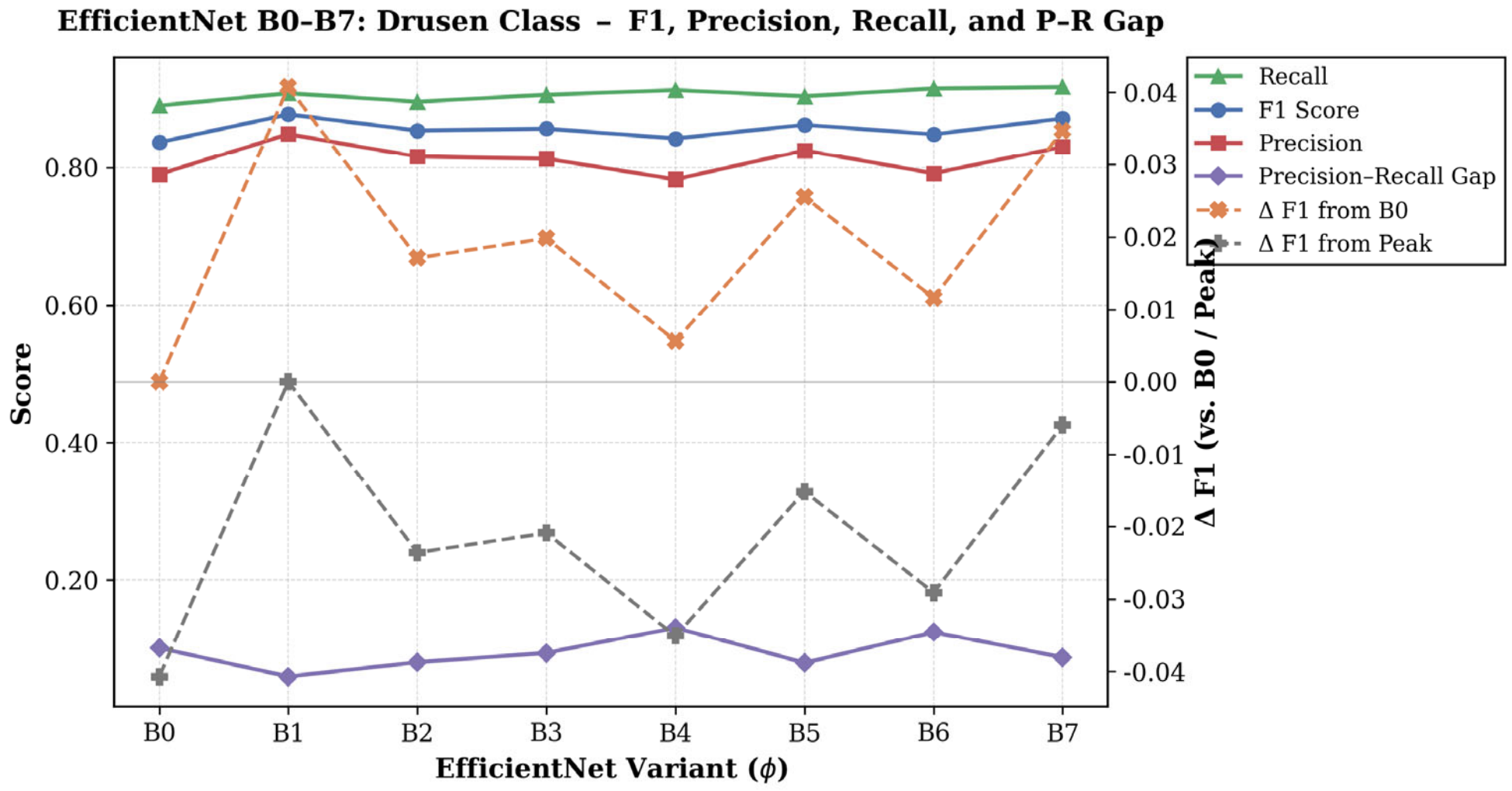

**Figure 10.** Drusen F1 Evolution Trajectory: B0 → B7.

The Drusen trajectory was 0.8373 for B0, 0.8780 for B1, 0.8544 for B2, 0.8571 for B3, 0.8429 for B4, 0.8628 for B5, 0.8489 for B6, and 0.8720 for B7. B1 achieved the highest observed Drusen F1 (0.8780), an absolute increase of 0.0407 relative to B0. B7 achieved the second-highest value (0.8720), which was 0.0060 below B1. B4 was the lowest among the higher-scale variants B1-B7 at 0.8429.

The class-specific response is further characterized by the distribution of F1 variability. Across the eight variants, Drusen accounted for 82.2% of the observed within-family variability in class-specific F1, quantified by the summed squared deviations described in Section 3.8.3. The corresponding contributions were 4.8% for CNV, 11.0% for DME, and 2.0% for Normal. These percentages were calculated from the class-specific F1 values reported in Table 9.

**Table 10.** Drusen F1 Evolution Trajectory: B0 → B7

| Model | Drusen F1 | Δ from B0 | Δ from peak | Precision | Recall | P-R Gap |
|---|---|---|---|---|---|---|
| B0 | 0.8373 | - | -0.0407 | 0.7900 | 0.8906 | 0.1006 |
| B1 | 0.8780 | 0.0407 | peak | 0.8493 | 0.9087 | 0.0594 |
| B2 | 0.8544 | 0.0171 | -0.0236 | 0.8162 | 0.8963 | 0.0801 |
| B3 | 0.8571 | 0.0198 | -0.0209 | 0.8129 | 0.9064 | 0.0935 |
| B4 | 0.8429 | 0.0056 | -0.0351 | 0.7826 | 0.9132 | 0.1306 |
| B5 | 0.8628 | 0.0255 | -0.0152 | 0.8251 | 0.9042 | 0.0791 |
| B6 | 0.8489 | 0.0116 | -0.0291 | 0.7914 | 0.9154 | 0.1240 |
| B7 | 0.8720 | 0.0347 | -0.0060 | 0.8306 | 0.9177 | 0.0871 |

The precision-recall decomposition provides additional detail on the Drusen trajectory. Table 11 shows that B1 achieved the highest Drusen F1 with precision of 0.8493 and recall of 0.9087. B4 and B6 achieved slightly higher recall than B1, at 0.9132 and 0.9154, respectively, but their precision decreased to 0.7826 and 0.7914. The precision-recall gap was smallest for B1 (0.0594), which also achieved the highest Drusen F1 (0.8780), whereas B4 and B6 exhibited larger gaps and lower Drusen F1 despite higher recall. For the one-vs-rest Drusen analysis, sensitivity and recall are equivalent and are therefore reported once as recall (sensitivity).

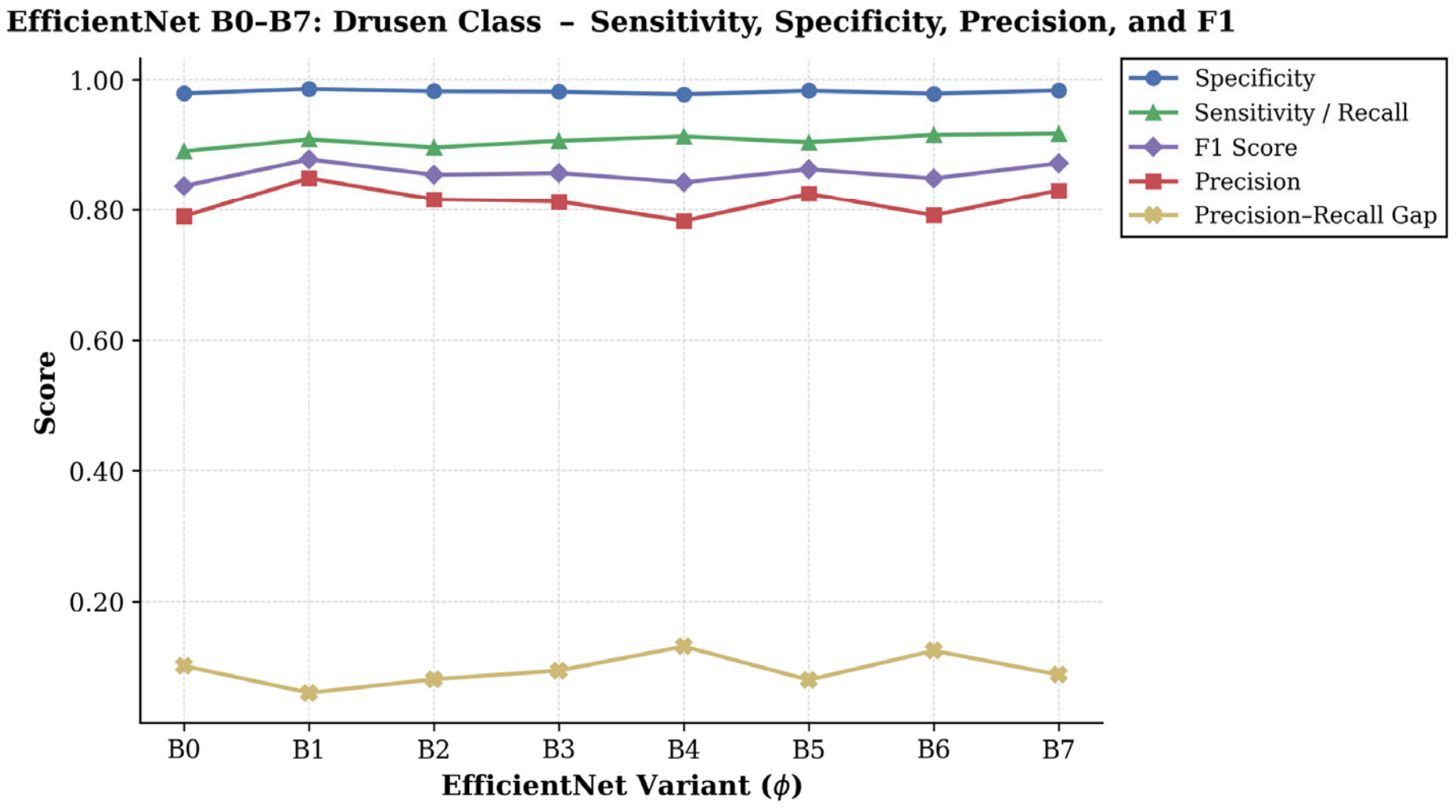

**Figure 11.** Drusen Specificity, Precision, Recall (Sensitivity), Precision-Recall Gap, and F1 Across EfficientNet-B0 → B7.

**Table 11.** Drusen Specificity, Precision, Recall (Sensitivity), Precision-Recall Gap, and F1 Across B0 → B7

| Model | Specificity | Precision | Recall(Sensitivity) | P-R Gap | Drusen F1 |
|---|---|---|---|---|---|
| B0 | 0.9791 | 0.7900 | 0.8906 | 0.1006 | 0.8373 |
| B1 | 0.9858 | 0.8493 | 0.9087 | 0.0594 | 0.8780 |
| B2 | 0.9822 | 0.8162 | 0.8963 | 0.0801 | 0.8544 |
| B3 | 0.9816 | 0.8129 | 0.9064 | 0.0935 | 0.8571 |
| B4 | 0.9776 | 0.7826 | 0.9132 | 0.1306 | 0.8429 |
| B5 | 0.9831 | 0.8251 | 0.9042 | 0.0791 | 0.8628 |
| B6 | 0.9787 | 0.7914 | 0.9154 | 0.1240 | 0.8489 |
| B7 | 0.9835 | 0.8306 | 0.9177 | 0.0871 | 0.8720 |

The F1 ranking was identical across all eight variants: Normal > CNV > DME > Drusen.

The difference between the highest and lowest class F1 values, denoted as $\Delta$, varied across the scaling spectrum. B1 produced the smallest observed disparity ($\Delta$ = 0.1058), whereas B0 produced the largest ($\Delta$ = 0.1434). B4 showed the second-largest disparity at 0.1360, while B7 yielded $\Delta$ = 0.1124.

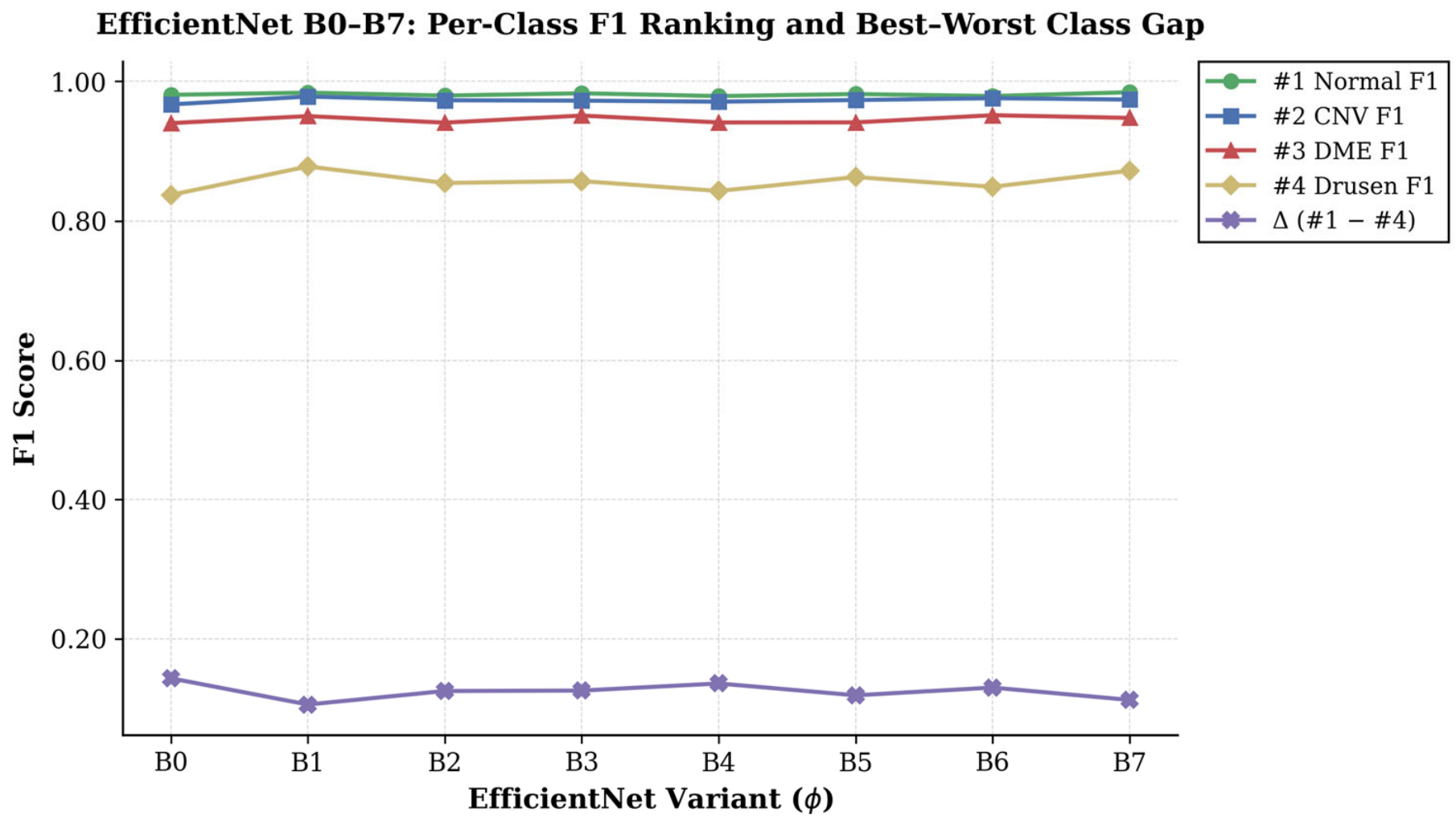

**Figure 12.** Class Ranking by F1 (Highest → Lowest per model).

B1 simultaneously achieved the highest test accuracy, the highest Drusen F1, and the smallest observed best-to-worst class-F1 difference ($\Delta$ = 0.1058).

**Table 12.** Class Ranking by F1 (Highest → Lowest per model)

| Model | #1(Normal) | #2(CNV) | #3(DME) | #4(Drusen) | Δ (#1-#4) |
|---|---|---|---|---|---|
| B0 | 0.9807 | 0.9669 | 0.9401 | 0.8373 | 0.1434 |
| B1 | 0.9838 | 0.9783 | 0.9502 | 0.8780 | 0.1058 |
| B2 | 0.9796 | 0.9729 | 0.9408 | 0.8544 | 0.1252 |
| B3 | 0.9829 | 0.9724 | 0.9510 | 0.8571 | 0.1258 |
| B4 | 0.9789 | 0.9709 | 0.9411 | 0.8429 | 0.1360 |
| B5 | 0.9818 | 0.9732 | 0.9412 | 0.8628 | 0.1190 |
| B6 | 0.9790 | 0.9759 | 0.9515 | 0.8489 | 0.1301 |
| B7 | 0.9844 | 0.9739 | 0.9476 | 0.8720 | 0.1124 |

### *4.3. ROC and Precision-Recall Analysis*

Figure 13 presents the one-vs-rest receiver operating characteristic (ROC) and precision-recall (PR) curves for all eight EfficientNet variants. Across the eight variants, the ROC curves were concentrated near the upper-left region for all four classes, indicating consistently high one-vs-rest discrimination across the evaluated test set. In contrast, the PR curves show greater class dependence, with CNV and Normal remaining close to the upper boundary, DME showing intermediate behavior, and Drusen consistently exhibiting the weakest precision-recall characteristics.

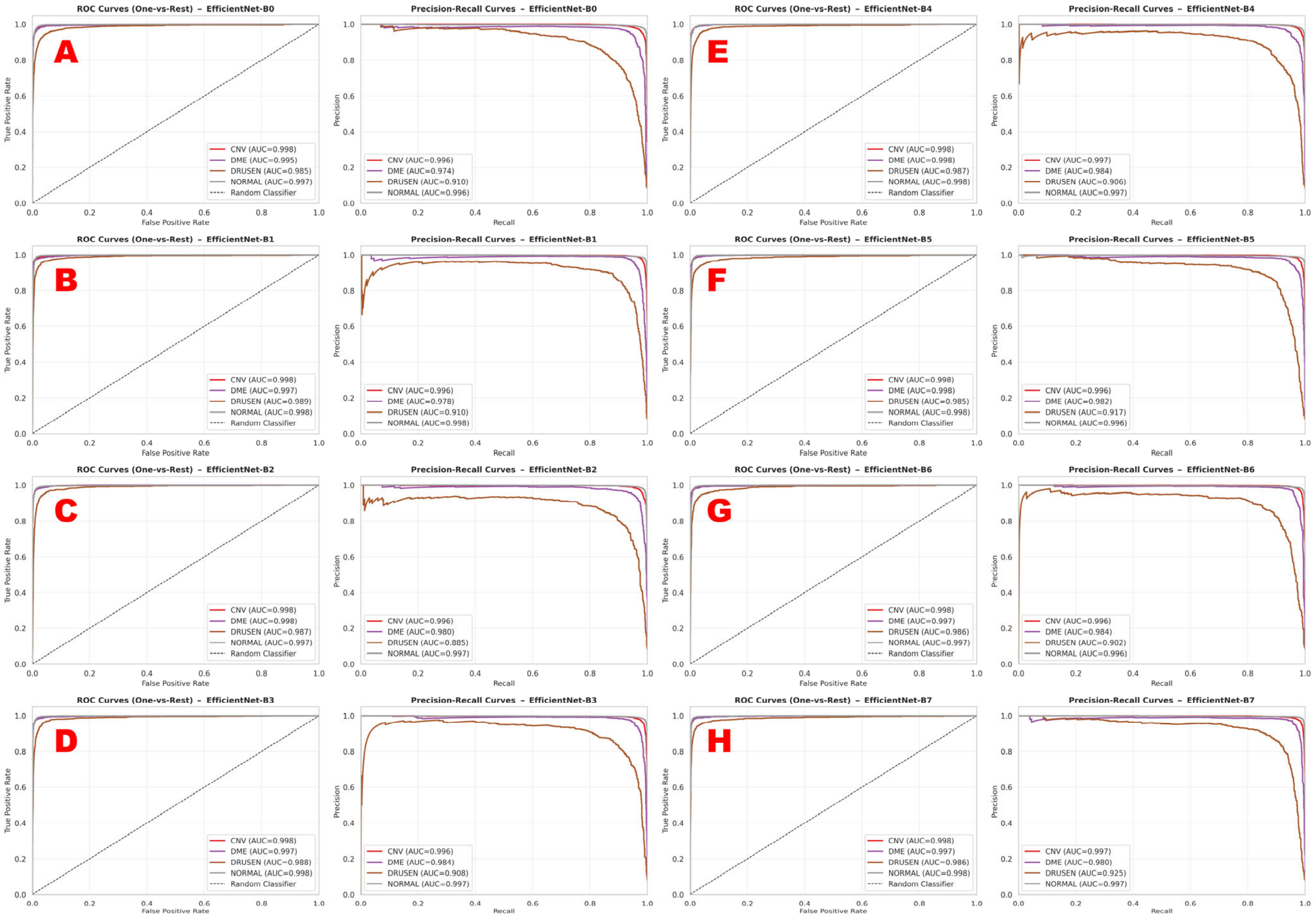

**Figure 13.** One-vs-rest receiver operating characteristic (ROC) curves across the eight evaluated backbones. A) EfficientNet-B0; B) EfficientNet-B1; C) EfficientNet-B2; D) EfficientNet-B3; E) EfficientNet-B4; F) EfficientNet-B5; G) EfficientNet-B6; H) EfficientNet-B7.

The contrast between ROC and PR analyses is most pronounced for Drusen: ROC-AUC remains within a narrow high range, whereas PR-AUC varies from 0.885 to 0.925 across the family. The difference between the ROC and PR views is most evident for Drusen. Although recall increases from 0.8906 at B0 to 0.9177 at B7, precision varies non-monotonically, reaching 0.8493 at B1 and decreasing to 0.7826 at B4. Thus, the higher recall achieved by several larger variants does not correspond to higher Drusen F1.

The PR curves provide complementary threshold-dependent information, while PR-AUC summarizes the corresponding precision-recall performance across thresholds of the class-wise behavior presented in Section 4.2. Intermediate variants showed greater recall without a corresponding precision increase, while B1 exhibited the smallest precision-recall gap (0.0594), whereas B4 and B6 showed larger gaps despite higher Drusen recall. B7 partially reduced the gap relative to B4 and B6. The ROC and precision-recall analyses therefore show that similar threshold-independent discrimination can coexist with greater variation in class-specific precision-recall characteristics, particularly for Drusen.

Importantly, the present findings do not indicate a progressive deterioration in threshold-independent discrimination across the EfficientNet family. Rather, the scaling-dependent variation is more apparent in the precision-recall behavior of the Drusen class, which exhibited the largest observed PR-AUC variation.

### *4.4. Confusion matrix structure*

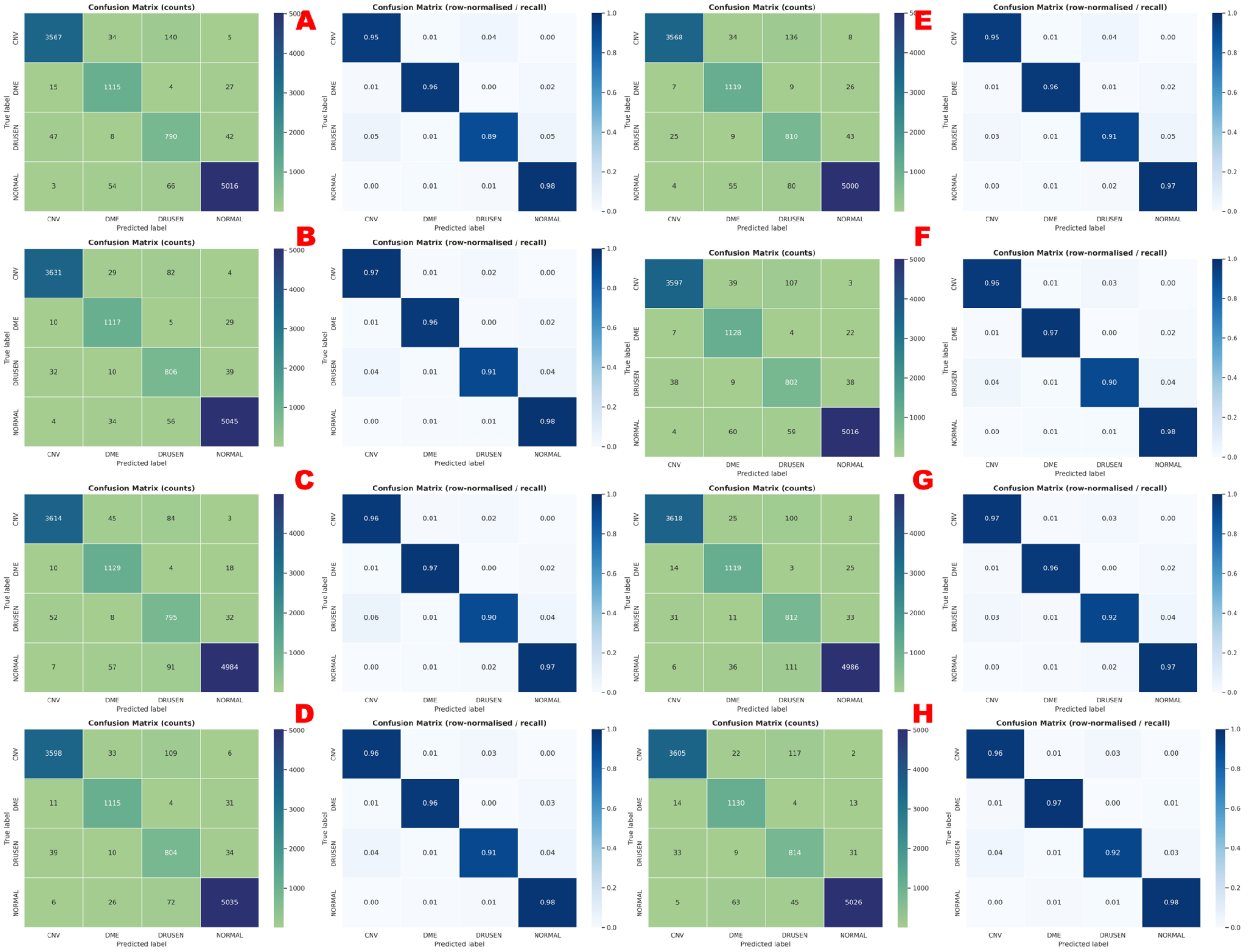

**Figure 14.** Confusion Matrices across eight backbones. A) EfficientNet-B0; B) EfficientNet-B1; C) EfficientNet-B2; D) EfficientNet-B3; E) EfficientNet-B4; F) EfficientNet-B5; G) EfficientNet-B6; H) EfficientNet-B7.

Figure 14 presents the confusion matrices for all eight EfficientNet variants. Across the scaling spectrum, the matrices retain a strong diagonal structure, indicating generally high class-level recognition for CNV, DME, Drusen, and Normal. The principal between-variant differences were concentrated in errors involving Drusen rather than in a uniform degradation of four-class recognition.

The principal between-variant differences in the confusion matrices involved Drusen-related errors. Compared with B1, B4 and B6 showed lower Drusen precision (0.7826 and 0.7914 versus 0.8493), despite higher recall for both variants. This pattern is consistent with more non-Drusen images being assigned to the Drusen class and illustrates the distinction between increased sensitivity and improved overall class-level F1. B7 partially recovered the precision-recall balance, achieving the highest Drusen recall (0.9177) and the second-highest Drusen F1 (0.8720), although B1 retained the highest Drusen F1 (0.8780).

Taken together, the confusion-matrix results complement the F1 and precision-recall analyses by showing that the scaling-dependent variation is concentrated primarily in Drusen-related misclassification patterns rather than representing a global loss of multiclass recognition.

### *4.5. Training Dynamics and Validation Performance*

Validation accuracy increased from Stage 1 to Stage 2 for all eight variants. Stage 2 validation accuracy ranged from 0.9623 for B0 to 0.9701 for B5 and B7. Table 13 reports Stage 1 and Stage 2 validation and training accuracies together with the final train-test gap.

**Table 13.** Stage 1 / Stage 2 Training Progression (Validation Accuracy/Training Accuracy)

| Model | Stage 1 Val. Acc. | Stage 1Train. Acc. | Stage 2 Val. Acc. | Stage 2Train. Acc. | Final Train-Test Gap (pp) |
|---|---|---|---|---|---|
| B0 | 0.8764 | 0.8517 | 0.9623 | 0.9646 | 0.53pp |
| B1 | 0.8207 | 0.8138 | 0.9690 | 0.9808 | 1.13pp |
| B2 | 0.8607 | 0.8327 | 0.9637 | 0.9707 | 0.83pp |
| B3 | 0.8663 | 0.8328 | 0.9682 | 0.9801 | 1.49pp |
| B4 | 0.8612 | 0.8097 | 0.9636 | 0.9781 | 1.80pp |
| B5 | 0.8875 | 0.8205 | 0.9701 | 0.9782 | 1.39pp |
| B6 | 0.8590 | 0.8098 | 0.9700 | 0.9816 | 1.80pp |
| B7 | 0.8732 | 0.8322 | 0.9701 | 0.9847 | 1.74pp |

* The final train-test gap is computed as final training accuracy minus held-out test accuracy. Stage 1 and Stage 2 validation accuracies are reported descriptively from the fixed validation split. Validation loss (val_loss) was used as the monitoring metric for both EarlyStopping and ModelCheckpoint.

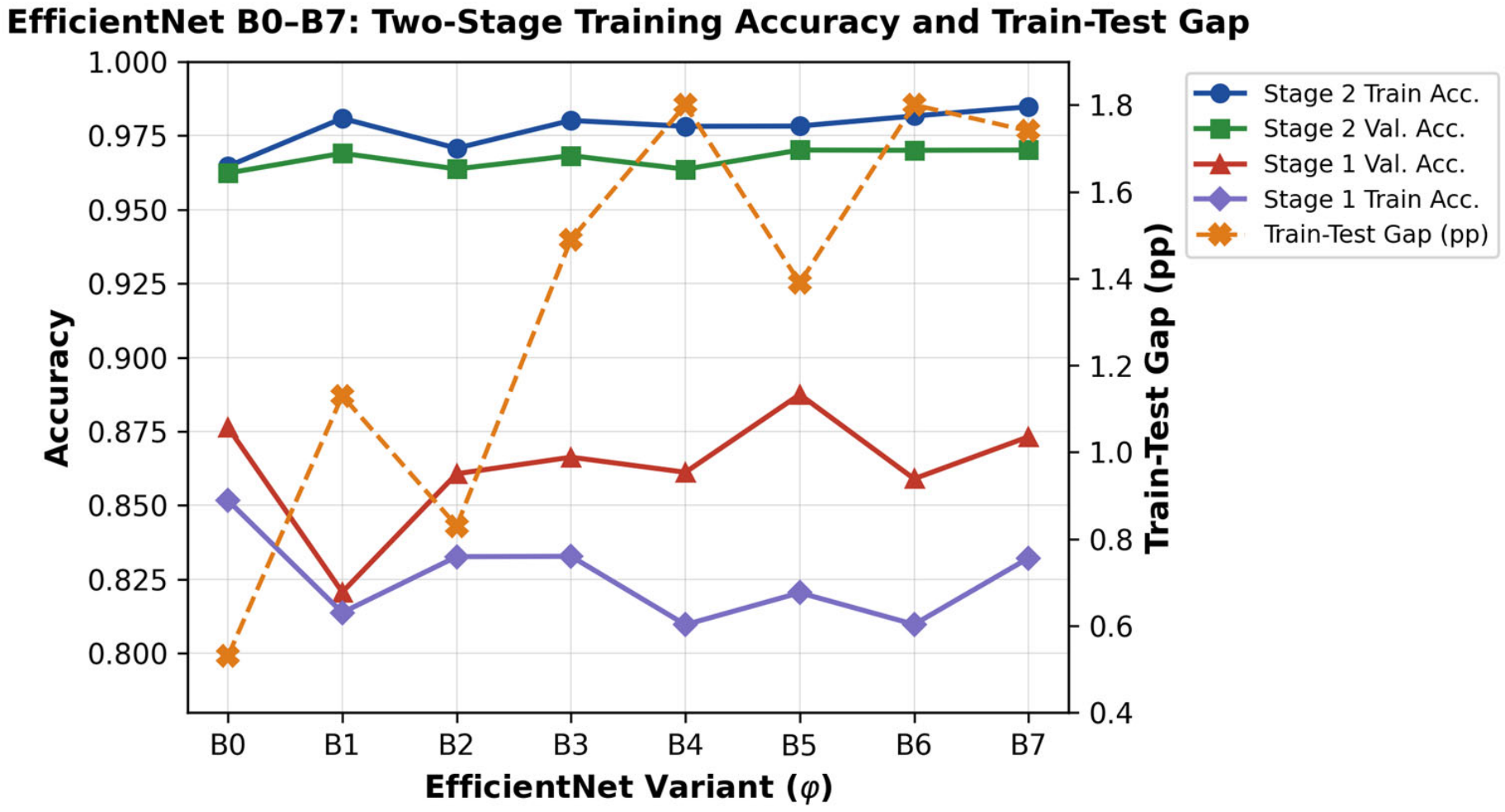

**Figure 15.** Stage 1 / Stage 2 Training Progression (Validation Accuracy/Training Accuracy).

The final train-test gap varied from 0.53 percentage points for B0 to 1.80 percentage points for B4 and B6. Its Pearson correlation with test accuracy was weak and non-significant (r = 0.26, p = 0.53), providing no evidence within this eight-model sample that the observed accuracy trajectory was explained by this gap.

The corresponding training dynamics are illustrated in Figures 15 and 16. All models demonstrate a substantial transition in validation performance from Stage 1 to Stage 2. This increase coincided with the transition from frozen-backbone training to partial backbone unfreezing and joint fine-tuning described in the experimental protocol. The magnitude of the resulting train-test gap, however, differs across variants.

The train-test gap should therefore be interpreted as a descriptive generalization measure rather than as evidence of a causal overfitting mechanism. In particular, the weak and non-significant correlation between the gap and test accuracy reported in Section 4.1 does not support attributing the non-monotonic scaling trajectory to the train-test gap alone.

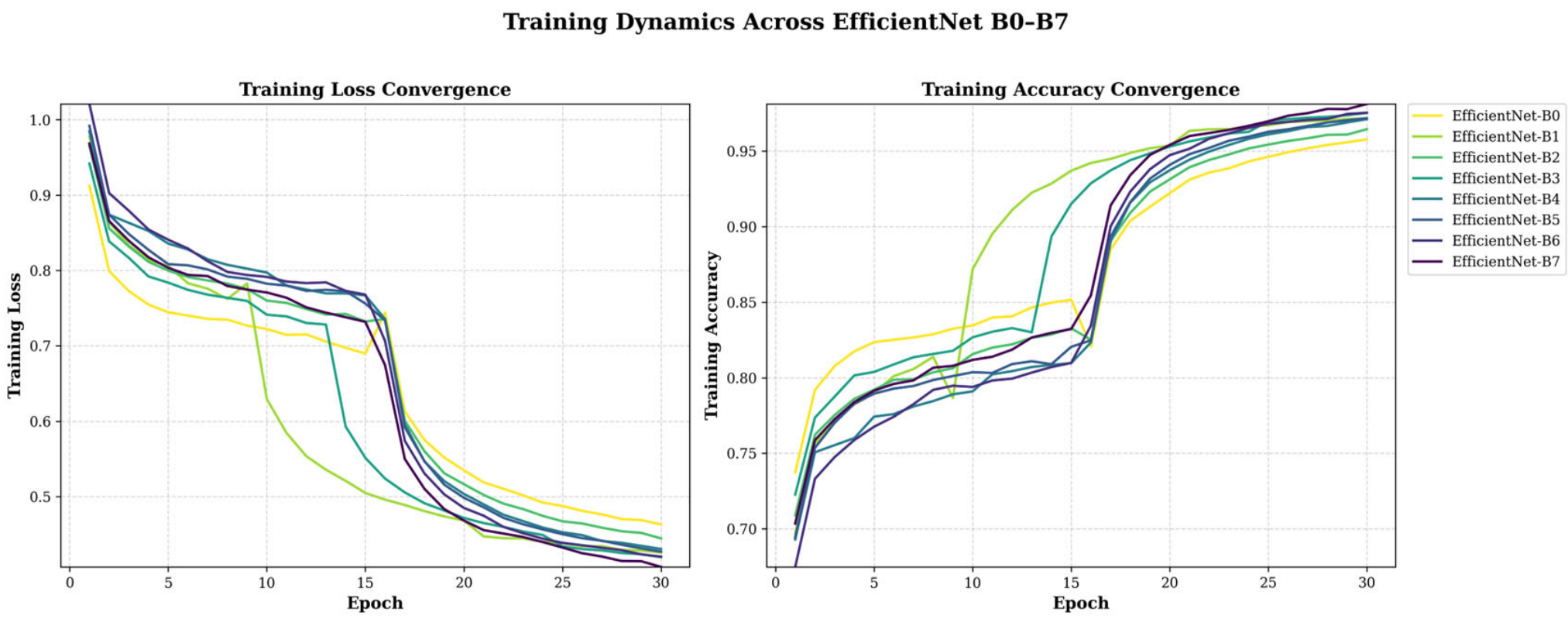

**Figure 16.** Training Dynamics Across EfficientNet-B0 → B7.

## *4.6. Inference Latency and Accuracy-to-Parameter Trade-off*

Table 14 summarizes model size, measured mean CPU inference latency, test accuracy, Drusen F1, and the accuracy-per-parameter ratio. Backbone and total parameter count increased monotonically from B0 to B7, while mean CPU inference latency increased substantially overall, from 407.48 ms for B0 to 1247.88 ms for B7. However, latency was not strictly monotonic across every adjacent variant, with B2 exhibiting a lower mean latency than B1. Test accuracy likewise did not increase monotonically.

**Table 14.** Inference Latency and Descriptive Accuracy-to-Parameter Ratio, B0 → B7.

| Model | Backbone Params | Total Params | Mean CPU Latency (ms) | Accuracy | Drusen F1 | Accuracy/Param (%/M) |
|---|---|---|---|---|---|---|
| B0 | 4.05M | 4.84M | 407.48 | 95.93% | 0.8373 | 19.82 |
| B1 | 6.58M | 7.37M | 512.79 | 96.95% | 0.8780 | 13.15 |
| B2 | 7.77M | 8.63M | 479.46 | 96.24% | 0.8544 | 11.15 |
| B3 | 10.78M | 11.71M | 544.27 | 96.52% | 0.8571 | 8.24 |
| B4 | 17.67M | 18.73M | 683.46 | 96.01% | 0.8429 | 5.13 |
| B5 | 28.51M | 29.70M | 844.98 | 96.43% | 0.8628 | 3.25 |
| B6 | 40.96M | 42.28M | 971.32 | 96.36% | 0.8489 | 2.28 |
| B7 | 64.10M | 65.54M | 1247.88 | 96.73% | 0.8720 | 1.48 |

The latency measurements showed a substantial increase in CPU inference cost across the EfficientNet family. Mean batch-1 latency increased from 407.48 ms for B0 to 1247.88 ms for B7, corresponding to a 3.06-fold increase, while throughput decreased from 2.454 to 0.801 images/s. The overall trend was consistent with increasing model size, although it was not strictly monotonic at every adjacent step because B2 showed lower mean latency than B1 (479.46 vs. 512.79 ms). Thus, model scale substantially increased measured CPU inference cost, whereas test accuracy remained non-monotonic. B1 is notable as the observed accuracy-cost operating point: it achieved the highest test accuracy and Drusen F1 while requiring 7.37M total parameters and 512.79 ms mean CPU latency, compared with 65.54M parameters and 1247.88 ms for B7.

The descriptive accuracy-to-parameter ratio decreased from 19.82 %/M for B0 to 13.15 %/M for B1 and 1.48 %/M for B7, corresponding to an approximately 13.4-fold reduction from B0 to B7. Despite this reduction, B7 achieved 96.73% test accuracy, only 0.22 percentage points below B1 (96.95%).

Relative to B0, B1 increases backbone parameters by approximately 1.63× while improving accuracy by 1.02 percentage points. B7 increases backbone parameters by approximately 15.83× relative to B0 while improving accuracy by only 0.80 percentage points. Relative to B0, the incremental test-accuracy gain per additional million backbone parameters were approximately 0.41 percentage points/M for B1 and 0.013 percentage points/M for B7.

### *4.7. Predictive-Probability Structure*

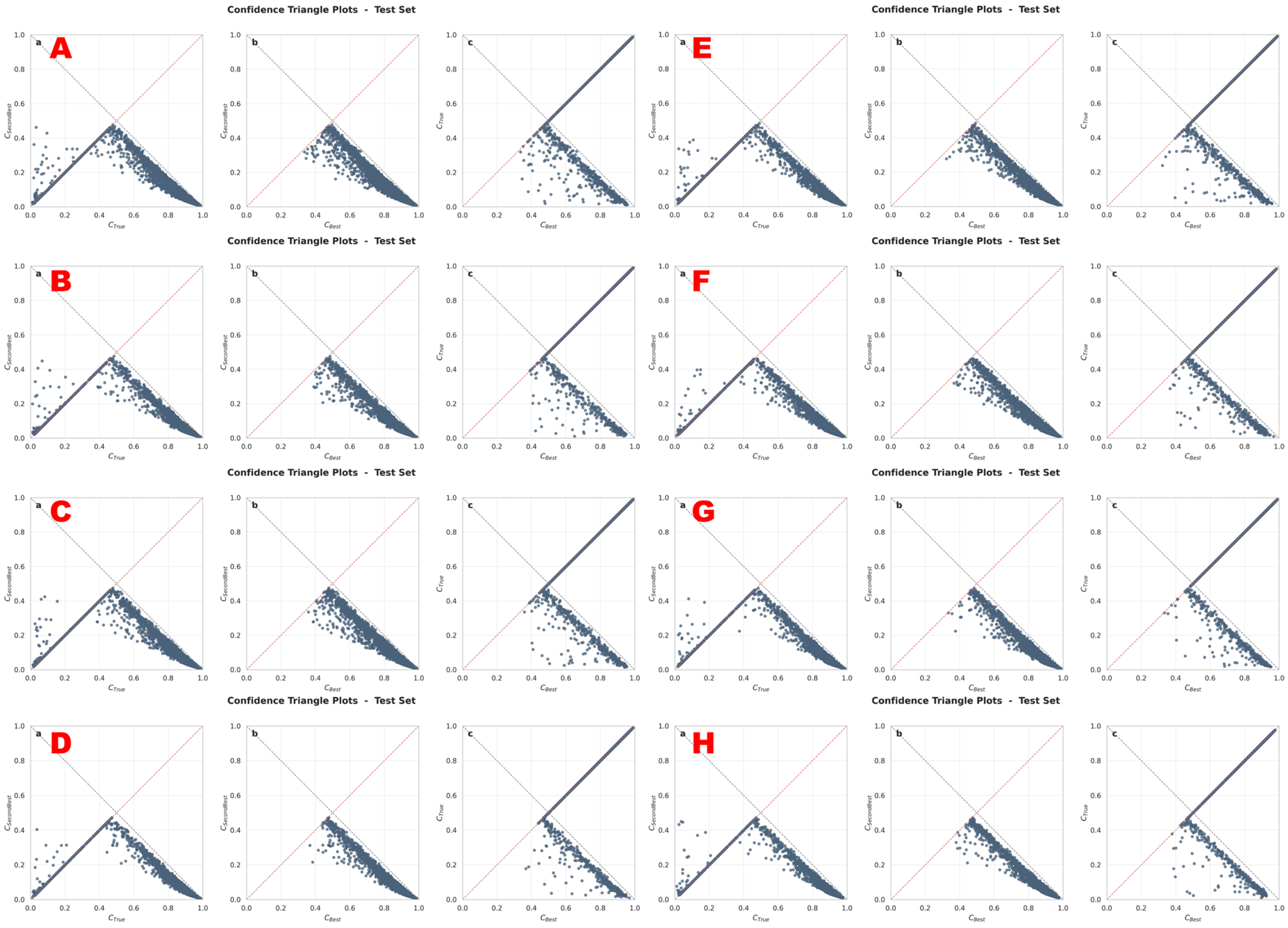

**Figure 17.** Predictive probability structure across the eight evaluated backbones. A) EfficientNet-B0; B) EfficientNet-B1; C) EfficientNet-B2; D) EfficientNet-B3; E) EfficientNet-B4; F) EfficientNet-B5; G) EfficientNet-B6; H) EfficientNet-B7.

Figure 17 presents the predictive-probability structure of the eight EfficientNet variants on the held-out test set using confidence-triangle plots. This analysis is qualitative and is intended to characterize the organization of predictive probabilities rather than to quantify probabilistic calibration.

Across B0 → B7, the predictive distributions exhibit a broadly stable triangular organization, with dense concentrations in higher-confidence regions and greater dispersion among ambiguous samples. The confidence-triangle plots showed broadly similar global probability organization across B0 → B7, with the principal visual differences occurring in lower-confidence regions.

The primary differences between variants occur in the distribution of lower-confidence predictions rather than in a wholesale transformation of the global probability geometry. High-confidence predictions remain concentrated around the comparatively well-separated diagnostic categories, while greater ambiguity is observed in regions exhibiting greater class overlap or predictive uncertainty. Greater dispersion of lower-confidence predictions was visually apparent in regions of greater class ambiguity, particularly for predictions involving Drusen.

The probability structure also does not exhibit a monotonic progression toward increasingly decisive predictions with increasing model capacity. B1 and B7, despite their large difference in parameter count, show broadly similar global probability organization and closely matched test accuracies of 96.95% and 96.73%, respectively. Intermediate variants likewise do not exhibit a qualitatively distinct probability geometry that would account for their lower aggregate performance solely through a global increase in predictive uncertainty.

The probability plots provide a qualitative complement to the ROC/PR and confusion-matrix analyses: the global confidence structure remains stable, but ambiguity in lower-confidence prediction regions is not systematically reduced by increasing architectural scale.

Because no numerical calibration metric was reported, the confidence-triangle plots should not be interpreted as quantitative evidence of improved or degraded calibration. Differences in visual concentration or dispersion of probabilities indicate qualitative changes in predictive structure rather than calibrated confidence performance.

### *4.8. Spatial Attribution and Feature Responses*

Figure 18 presents Grad-CAM visualizations for all eight EfficientNet variants, while Figure 19 shows representative intermediate feature maps across the same scaling spectrum. In the representative correctly classified examples, Grad-CAM activation was predominantly concentrated in central retinal regions, with comparatively limited activation in peripheral background regions.

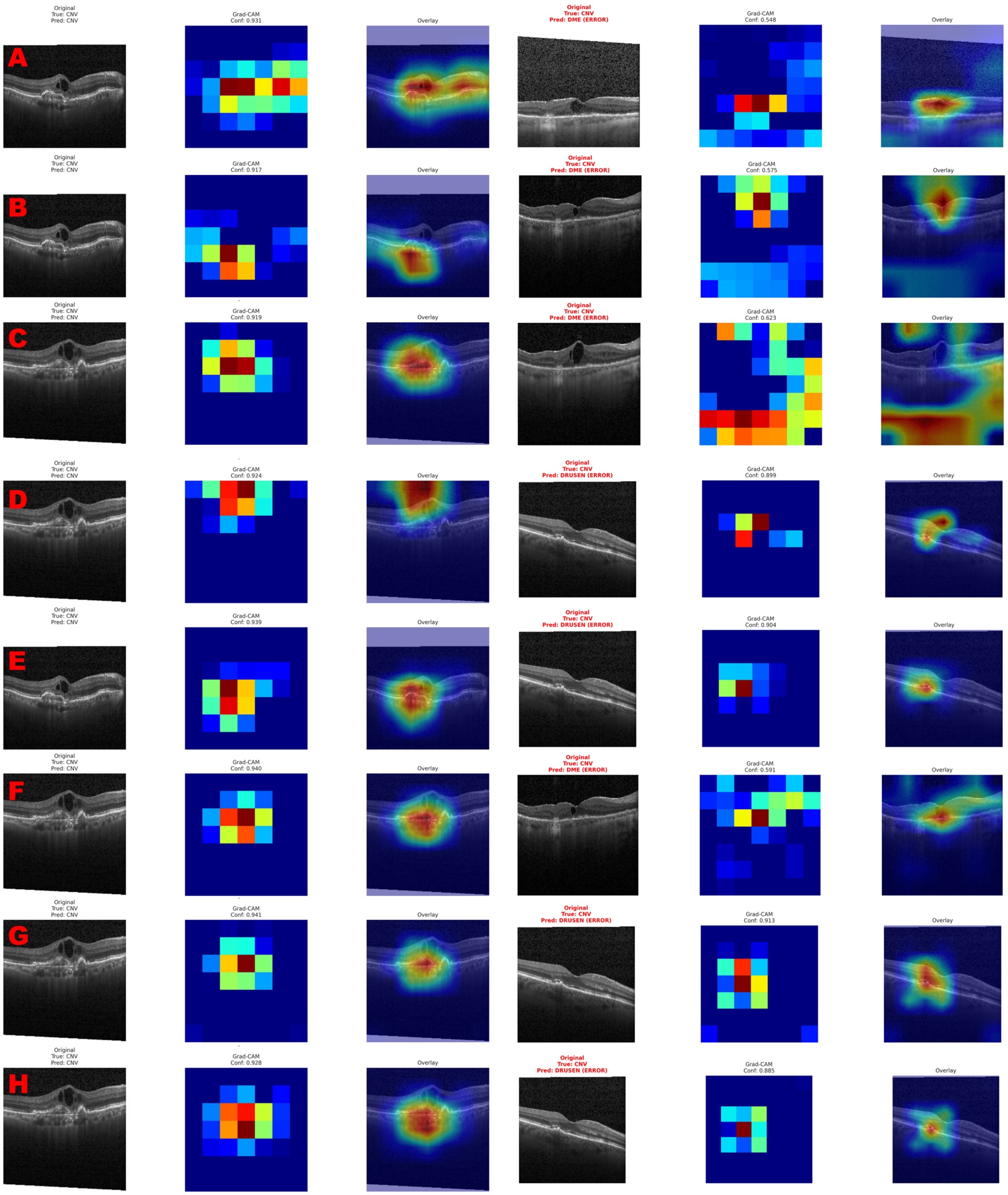

**Figure 18.** Grad-CAM across the eight evaluated backbones. A) EfficientNet-B0; B) EfficientNet-B1; C) EfficientNet-B2; D) EfficientNet-B3; E) EfficientNet-B4; F) EfficientNet-B5; G) EfficientNet-B6; H) EfficientNet-B7.

The attribution maps also show a distinction between correct and erroneous predictions. Representative CNV cases exhibit concentrated responses over central lesion-bearing retinal regions across the scaling spectrum. In contrast, representative Drusen cases misclassified as CNV exhibit substantial overlap in spatial attribution with the same central retinal structures. In the representative misclassified Drusen-as-CNV cases, the attribution maps remained concentrated in central retinal regions rather than peripheral background areas.

The representative Grad-CAM maps showed broadly similar spatial localization across several EfficientNet variants.

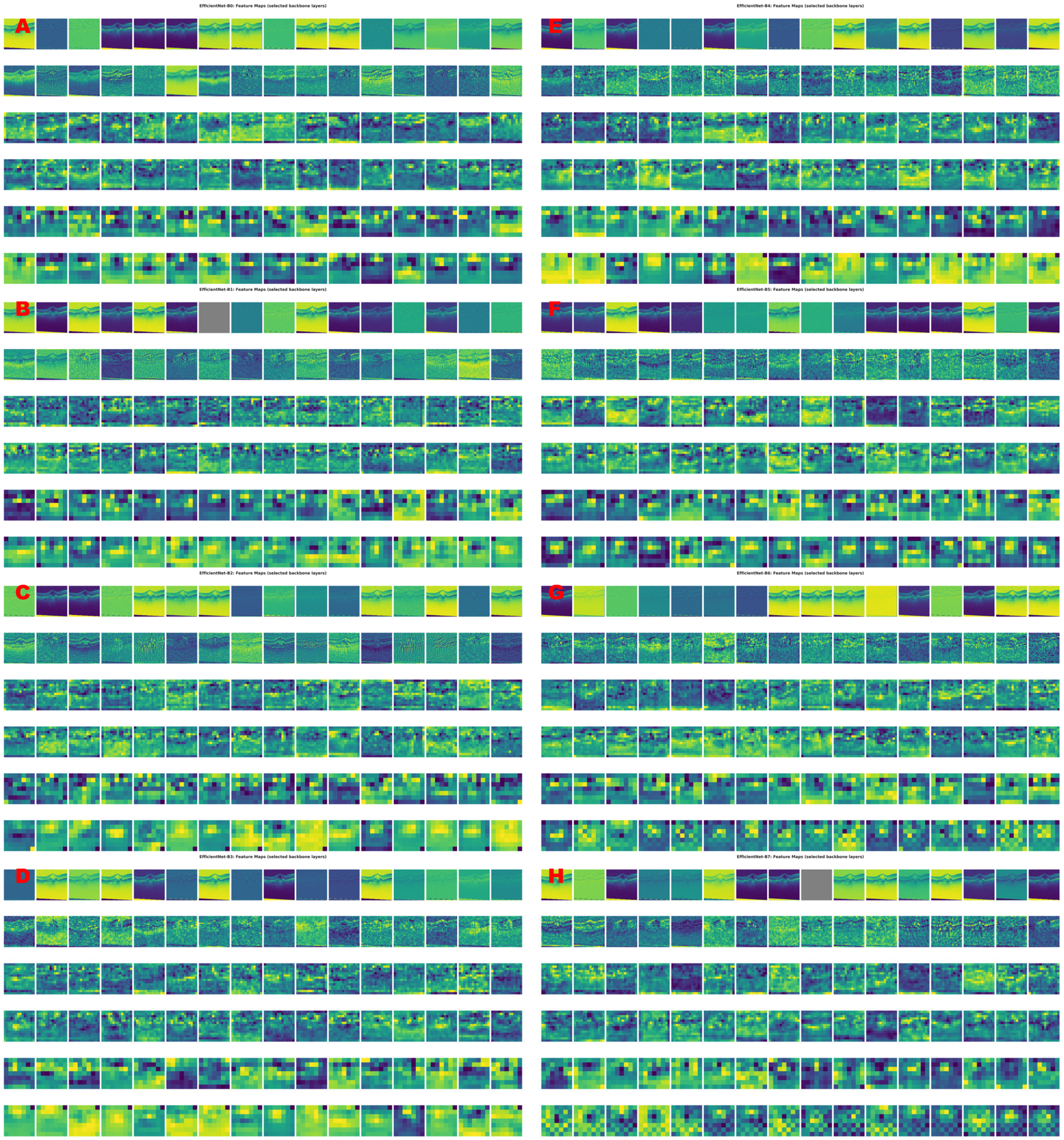

**Figure 19.** Feature maps across the eight evaluated backbones. A) EfficientNet-B0; B) EfficientNet-B1; C) EfficientNet-B2; D) EfficientNet-B3; E) EfficientNet-B4; F) EfficientNet-B5; G) EfficientNet-B6; H) EfficientNet-B7.

The intermediate feature maps provide a complementary representation-level view. The representative feature maps displayed increasingly heterogeneous activation patterns among the larger variants. However, this increase in feature-map complexity is not accompanied by a monotonic improvement in diagnostic performance. B1 achieved the highest test accuracy and Drusen F1 despite having fewer backbone parameters than B4–B7, whereas B7 showed only partial recovery relative to B1.

Taken together, the attribution and feature-response analyses indicate that the scaling-dependent differences are not characterized by a simple transition from irrelevant to relevant image attention. Instead, the variants generally exhibit spatial responses concentrated over retinal structures, while the main performance differences remain associated with how those responses relate to classification of classes exhibiting greater observed error and inter-class overlap.

### *4.9. Representation Geometry*

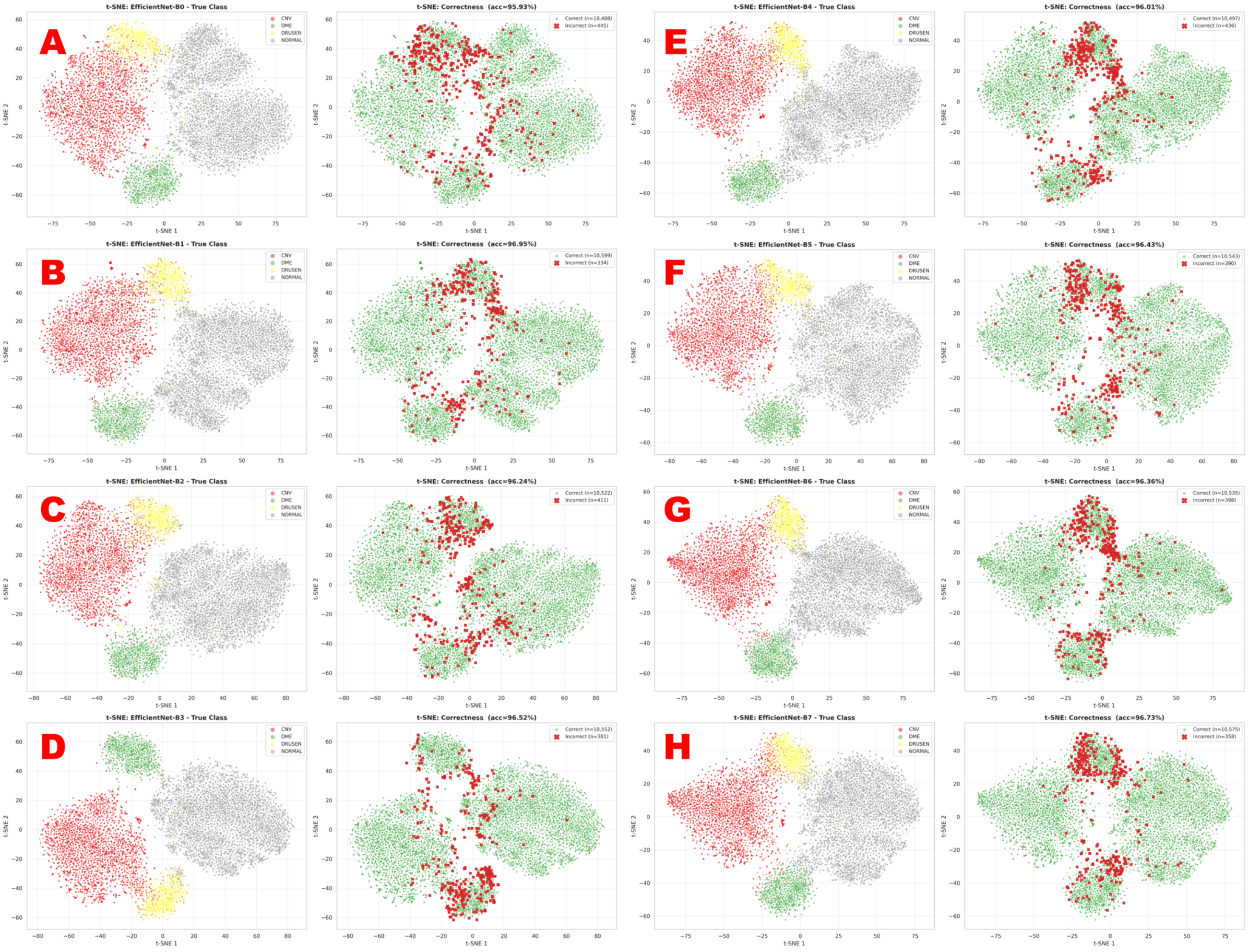

**Figure 20.** t-SNE across the eight evaluated backbones. A) EfficientNet-B0; B) EfficientNet-B1; C) EfficientNet-B2; D) EfficientNet-B3; E) EfficientNet-B4; F) EfficientNet-B5; G) EfficientNet-B6; H) EfficientNet-B7.

Figure 20 presents t-SNE projections of the learned representations for all eight EfficientNet variants, visualized both by true diagnostic class and by prediction correctness. Across B0 → B7, the t-SNE projections showed broadly similar qualitative class organization, with CNV and Normal appearing more separated than DME and Drusen. CNV and Normal form comparatively distinct regions, whereas DME and Drusen occupy smaller regions with greater local overlap and inter-class mixing.

The correctness-colored projections provided a qualitative view of the locations of correctly and incorrectly classified samples within the projected representation space. Correctly classified samples dominate the principal class regions, whereas incorrect predictions were visually concentrated in transition zones between overlapping class distributions. The most visible mixing occurs among the non-Normal classes, consistent with the lower F1 values observed for Drusen across all variants.

No monotonic increase in apparent cluster compactness or separation was observed across the t-SNE projections as model size increased. In particular, B1 and B7 exhibit qualitatively similar global class topology despite their substantial difference in parameter count and their closely matched test accuracies of 96.95% and 96.73%, respectively.

The representation analysis should nevertheless be interpreted qualitatively rather than as a quantitative measure of embedding separability. Because t-SNE is a nonlinear dimensionality-reduction procedure, local neighborhood structure is more informative than absolute distances between clusters; accordingly, the two-dimensional projections should not be interpreted as direct quantitative measurements of class separability.

### *4.10. Integrated Summary of Observed Scaling Findings*

Across the evaluated EfficientNet family, test accuracy and class-specific performance varied non-monotonically with model scale. B1 achieved the highest observed test accuracy (96.95%) and Drusen F1 (0.8780), whereas B7 achieved 96.73% test accuracy and 0.8720 Drusen F1 at substantially greater parameter count and measured CPU inference latency.

The class-specific analysis showed that Drusen remained the lowest-F1 class across all eight variants and accounted for the largest proportion of the summed within-family F1 variation. The precision-recall analysis further showed that the highest Drusen recall was achieved by B7, but this did not correspond to the highest Drusen F1 because precision remained lower than that of B1.

The ImageNet ranking was not reproduced in the OCT test-accuracy ranking. Within the eight evaluated variants, the source-domain ordering was not preserved in the observed OCT test-accuracy ranking, and the corresponding Spearman association was weak and non-significant.

The computational analysis showed that model size and measured CPU inference cost increased substantially across the evaluated family, whereas the corresponding diagnostic performance did not increase monotonically. The descriptive accuracy-to-parameter ratio also decreased markedly as model size increased.

Taken together, the Results demonstrate that, under the evaluated fixed-resolution transfer-learning protocol, increasing EfficientNet scale did not produce a monotonic improvement in four-class SD-OCT classification performance. The observed behavior is therefore characterized by a non-monotonic accuracy trajectory, class-dependent variation concentrated particularly in Drusen, and increasing computational cost with increasing model scale.

To summarize the principal observations without introducing additional interpretation, the original five finding categories are consolidated in Table 15.

**Table 15.** Integrated Summary of the Observed Scaling Findings.

| Finding | Empirical Statement |
|---|---|
| F1 | B1 ($\varphi = 1$) achieved the highest observed test accuracy (96.95%) |
| F2 | Test accuracy varied non-monotonically across B0 $\rightarrow$ B7 |
| F3 | Drusen F1 followed a non-monotonic trajectory, peaking at B1 and partially recovering at B7 |
| F4 | Within the eight evaluated variants, the ImageNet Top-1 ranking showed a weak and non-significant association with the OCT test-accuracy ranking |
| F5 | The descriptive accuracy-to-parameter ratio decreased substantially with increasing model size, from 19.82 %/M for B0 to 1.48 %/M for B7 |

## 5. Discussion

The present study evaluated the complete EfficientNet-B0 → B7 family under a common fixed-resolution transfer-learning protocol for four-class SD-OCT retinal disease classification. The principal finding was that increasing model scale did not produce a monotonic improvement in downstream classification performance. Test accuracy increased from 95.93% for B0 to 96.95% for B1, subsequently fluctuated across the intermediate variants, and recovered to 96.73% for B7. A similar non-monotonic trajectory was observed for Drusen F1, which peaked at B1 and partially recovered at B7. These findings indicate that the relationship between EfficientNet scale and downstream performance is not necessarily preserved after transfer from ImageNet to the evaluated OCT task.

### *5.1. Non-Monotonic Transfer of EfficientNet Scaling to SD-OCT*

The most prominent observation is the difference between the source-domain EfficientNet hierarchy and the target-domain OCT hierarchy. The canonical ImageNet results increase progressively from B0 through B7, with B7 occupying the highest Top-1 accuracy position. In contrast, the present OCT test ranking was B1, B7, B3, B5, B6, B2, B4, and B0. B1 therefore achieved the highest observed OCT test accuracy despite occupying only the seventh position in the ImageNet Top-1 hierarchy.

The Spearman rank correlation between the ImageNet Top-1 ranking and the OCT test-accuracy ranking was weakly positive and non-significant ($\rho = 0.286$, $p = 0.493$). With only eight architectural variants, this result should be interpreted as an exploratory measure of ranking correspondence; the present data do not provide evidence of a statistically reliable association between the source-domain and target-domain rankings.

Several factors may contribute to the discrepancy. The ImageNet benchmark and the present OCT task differ substantially in image statistics, semantic structure, class cardinality, and target-domain representation requirements. More importantly, the present experiment fixes input resolution at $224 \times 224$. The resulting comparison therefore does not reproduce the complete EfficientNet depth-width-resolution scaling trajectory. It examines the response to increasing architectural scale under a constrained spatial-resolution condition. The observed target-domain trajectory consequently reflects the combined effect of the EfficientNet depth and width configurations, the fixed input resolution, ImageNet initialization, the adopted fine-tuning schedule, and the OCT data distribution. The experiment does not independently isolate these factors.

The B0-to-B1 transition is particularly informative. B1 increased backbone parameters from 4.05M to 6.58M, corresponding to an approximately 1.63-fold increase, while test accuracy increased by 1.02 percentage points. However, continued scaling did not preserve this gain. B7 contained 64.10M backbone parameters approximately 15.83 times the B0 backbone size but improved test accuracy relative to B0 by only 0.80 percentage points and remained 0.22 percentage points below B1.

This pattern should not be interpreted as evidence that larger EfficientNet models are intrinsically inferior for OCT. Rather, it demonstrates that, under the evaluated transfer-learning configuration, additional architectural capacity did not translate into a proportional or monotonic downstream benefit. The result is consistent with the broader transfer-learning literature indicating that the usefulness of a source-trained representation depends on the relationship between source and target domains and on the extent of adaptation required during fine-tuning.

### *5.2. Class-Specific Scaling Behavior and Drusen Performance*

The class-specific analyses provide a more informative view of the scaling response than aggregate accuracy alone. Drusen was the lowest-F1 class for every evaluated variant, whereas CNV and Normal exhibited relatively narrow F1 ranges. Drusen F1 varied from 0.8373 to 0.8780, compared with 0.9669-0.9783 for CNV and 0.9789-0.9844 for Normal.

The concentration of within-family F1 variability in Drusen is especially notable. Drusen accounted for 82.2% of the summed squared deviations of class-specific F1 values across the eight variants, compared with 4.8% for CNV, 11.0% for DME, and 2.0% for Normal. This calculation indicates that the largest proportion of the observed class-specific F1 variability across the evaluated variants was concentrated in Drusen.

This finding suggests that aggregate measures such as accuracy and weighted F1 can mask important class-specific effects. The overall test accuracy remained high across all variants, yet the largest variation among class-level F1 values occurred in the minority Drusen class. Accordingly, within-family scaling should not be evaluated solely through aggregate metrics when the objective is to understand how architectural capacity affects individual diagnostic categories.

The Drusen trajectory was also distinctly non-monotonic. B1 achieved the highest Drusen F1 (0.8780), B4 the lowest among B1-B7 (0.8429), and B7 partially recovered to 0.8720. Notably, B7 achieved the highest Drusen recall (0.9177), but its F1 remained below B1 because its precision was lower (0.8306 versus 0.8493). Similarly, B4 and B6 achieved recall values above that of B1 but substantially lower precision values.

These observations indicate that the effect of model scaling on Drusen performance is not adequately described by sensitivity or recall alone. Instead, the observed differences involve changes in the balance between precision and recall. The Pearson correlation between the Drusen precision-recall gap and Drusen F1 was negative ($r=-0.76$), which is consistent with the observed relationship across the eight variants. However, this correlation is descriptive and does not establish that precision causally determines F1. In addition, the experiment did not independently manipulate or optimize the Drusen decision threshold. The appropriate interpretation is therefore an observed precision-recall trade-off rather than a demonstrated causal mechanism.

The confusion matrices provide complementary evidence. The most visible between-variant differences were associated with errors involving Drusen, including both Drusen observations assigned to other classes and non-Drusen observations assigned to Drusen. B4, for example, showed lower Drusen precision than B1, while B7 partially restored the balance between precision and recall. These patterns are consistent with the class-specific metric analysis, but they do not by themselves establish whether the underlying differences arise from learned feature representations, classifier thresholds, or another component of the inference process.

### *5.3. Accuracy-Efficiency Trade-off*

The predictive results must be considered jointly with computational cost. B1 achieved the highest observed test accuracy (96.95%) and the highest Drusen F1 (0.8780) while requiring 7.37M total parameters and a mean CPU inference latency of 512.79 ms. B7 approached these predictive values, reaching 96.73% test accuracy and 0.8720 Drusen F1, but required 65.54M total parameters and 1247.88 ms mean CPU latency.

The difference becomes more pronounced when parameter-normalized performance is considered. The descriptive accuracy-to-parameter ratio decreased from 19.82 percentage points per million total parameters for B0 to 13.15 for B1 and 1.48 for B7. Thus, the increase in model size was accompanied by a substantial reduction in this parameter-normalized quantity. Relative to B0, B1 increased backbone parameters by approximately 1.63-fold while improving test accuracy by 1.02 percentage points. B7 increased backbone parameters by approximately 15.83-fold while improving accuracy by only 0.80 percentage points.

These results show a decreasing descriptive accuracy-to-parameter ratio across the evaluated variants as model size increases. They should not, however, be interpreted as establishing a general scaling law. The ratio used in this study is a descriptive metric, and it does not account for memory bandwidth, floating-point operations, energy consumption, accelerator-specific utilization, or other deployment variables.

The latency measurements reinforce the same practical distinction. Mean CPU latency increased from 407.48 ms for B0 to 1247.88 ms for B7, while throughput decreased from 2.454 to 0.801 images/s. The relationship was not strictly monotonic at every adjacent step, since B2 showed lower mean latency than B1, but the overall increase in computational cost with model size was substantial.

Within the evaluated configuration, B1 combined the highest observed test accuracy and Drusen F1 with substantially fewer total parameters and lower mean CPU inference latency than B7. This observation identifies B1 as an empirical operating point within the present protocol, rather than establishing it as a universally optimal EfficientNet configuration for OCT classification. The 0.22-percentage-point difference between B1 and B7 is small, particularly given that only a single nominal seed was evaluated. The data support selecting a model by jointly considering predictive performance and computational requirements rather than assuming that the largest available variant is preferable.

### *5.4. Training Dynamics and Progressive Fine-Tuning*

The recorded Stage 2 validation accuracy exceeded the corresponding Stage 1 validation accuracy for all eight variants. This behavior is consistent with the intended role of progressive fine-tuning: Stage 1 adapts the newly initialized classification head while keeping the pretrained backbone frozen, whereas Stage 2 permits adaptation of the upper portion of the backbone. Stage 2 validation accuracy ranged from 0.9623 for B0 to 0.9701 for B5 and B7.

However, the observed improvement across this transition should not be interpreted as evidence that the selected 50% unfreezing strategy is optimal. The study did not compare alternative freezing ratios, alternative fine-tuning schedules, or a single-stage baseline. Consequently, the contribution of the progressive fine-tuning design itself cannot be separated from the other components of the training protocol.

The final train-test gap ranged from 0.53 percentage points for B0 to 1.80 percentage points for B4 and B6. B1 achieved the highest test accuracy with an intermediate gap of 1.13 percentage points, while B0 had the smallest gap but the lowest test accuracy. The Pearson correlation between the final train-test gap and test accuracy was weak and non-significant ($\mathrm{r} = 0.26, p = 0.53$). Thus, the observed non-monotonic accuracy trajectory cannot be attributed to the final train-test gap alone.

This distinction is important because a larger train-test gap does not necessarily imply that a variant is responsible for the lower end-to-end performance observed at a particular scale. The gap is better interpreted here as a descriptive indicator of generalization behavior rather than as a direct measure of a specific overfitting mechanism.

The common seed of 42 provides a matched stochastic condition for the eight variants, but the single-seed design does not quantify between-run variability. This is particularly relevant because B1 and B7 differ by only 0.22 percentage points in test accuracy.

Independent repetitions across multiple seeds would be required to determine whether this ordering is robust to stochastic variation.

### *5.5. Predictive Probability, Spatial Attribution, and Representation Analysis*

The qualitative analyses provide complementary context but do not establish a mechanistic explanation for the non-monotonic scaling trajectory.

The confidence-triangle plots showed broadly similar global predictive-probability organization across B0 → B7, with greater visual dispersion in lower-confidence regions and in regions of greater class ambiguity. No monotonic progression toward increasingly decisive predictions was evident with increasing model size. Because no quantitative calibration metric was reported, these observations should not be interpreted as evidence that calibration either improved or deteriorated with scale.

The Grad-CAM analyses generally showed activation concentrated over central retinal structures rather than peripheral background regions. Representative Drusen-as-CNV errors also showed activation over central retinal structures, indicating that the misclassified examples were not simply associated with visually irrelevant background regions. Nevertheless, these representative maps are qualitative and do not establish whether the highlighted structures are sufficient or causally responsible for the classification outcome.

The intermediate feature maps displayed increasingly heterogeneous activation patterns among larger variants, but this visual complexity was not accompanied by monotonic improvement in classification performance. This finding is consistent with the central result that greater architectural capacity did not guarantee better downstream accuracy, although it does not establish why the additional features failed to translate into a consistent performance advantage.

The t-SNE projections provided a similar qualitative perspective. CNV and Normal occupied comparatively distinct regions, whereas DME and Drusen showed greater overlap and inter-class mixing. Incorrect predictions were visually more concentrated in regions of class overlap within the two-dimensional projections. No monotonic increase in apparent cluster compactness or separation was visible with increasing model size.

These observations should be interpreted cautiously because t-SNE is a nonlinear dimensionality-reduction method, and the two-dimensional geometry is not a direct quantitative measure of representation separability. The attribution and embedding analyses therefore support a qualitative description of the learned representations but should not be used to claim a specific representation-level mechanism for the superiority of B1 or the partial recovery observed at B7.

### *5.6. Limitations and Future Research*

Several limitations constrain the generalizability and interpretation of the present findings.

- **First,** the study uses a single publicly available OCT dataset. The observed non-monotonic scaling trajectory and Drusen-specific variation may therefore depend on the characteristics of this dataset, its acquisition devices, class composition, preprocessing, and predefined partitioning. External evaluation on independent datasets acquired using different devices and clinical workflows is required before the observed scaling behavior can be considered general across OCT populations.
- **Second,** all eight variants were evaluated using a single nominal random seed, seed = 42. Although the same seed establishes a matched stochastic condition, it does not quantify between-run variability. Repeated independent runs with multiple seeds would provide stronger evidence regarding the stability of the B1-versus-B7 ranking and the reproducibility of the observed class-specific trajectories.

- **Third,** the manuscript distinguishes canonical EfficientNet parameter counts from the experimentally instantiated counts produced by the implementation. The latter were used for the Results and efficiency analyses and therefore provide the appropriate basis for interpreting computational cost. Maintaining this distinction is necessary for avoiding apparent inconsistencies between the architectural specification table and the experimental measurements.
- **Fourth,** the same augmentation strategy was applied to every variant, which is appropriate for controlled architectural comparison but prevents separation of the individual contributions of the augmentation operations. Targeted ablation could determine whether the observed scaling behavior depends on specific geometric or photometric perturbations.
- **Fifth,** the experiment fixed input resolution at $224 \times 224$. It therefore does not reproduce the complete EfficientNet depth-width-resolution compound-scaling trajectory. The conclusions apply specifically to the evaluated fixed-resolution within-family comparison and should not be generalized to canonical EfficientNet scaling as a three-dimensional scaling law.
- **Sixth,** the inference measurements were obtained on a non-dedicated CPU system. They therefore characterize the relative computational burden under the specified hardware and software configuration rather than providing hardware-independent deployment benchmarks. Differences in processor architecture, optimized kernels, batch size, threading, quantization, accelerator availability, and runtime implementation could alter the absolute latency values.
- **Seventh**, the integrity and independence of the supplied dataset partitions have not been established. A published SHA-256 audit reported duplicate images across the training, validation, and test partitions of the OCT2018 variant with the same partition counts as those used here [52]. The specific files used in this study have not been independently audited for cross-partition duplicates. Consequently, test-set independence cannot be confirmed, and the reported metrics may not fully reflect performance on image content absent from training. A file-level integrity audit, followed where necessary by leakage-controlled retraining and evaluation, is required to establish the robustness of these results.

Several directions follow directly from these limitations. A multi-seed replication would provide uncertainty estimates for the relative ranking of B1 and B7. Evaluation on independent OCT datasets would test whether the non-monotonic trajectory generalizes beyond the Kermany benchmark. A factorial ablation of resolution, width, depth, fine-tuning depth, and data augmentation would help determine which components contribute most strongly to the observed downstream response. A broader cross-family benchmark involving ResNet, DenseNet, and lightweight architectures would additionally establish whether the observed intermediate-scale optimum is specific to EfficientNet or represents a more general property of transfer learning on this task.

An especially important extension would be to investigate the interaction between dataset size and model scale. The present experiment cannot determine whether larger variants would become preferable with substantially more target-domain training data. Similarly, the contribution of the SE mechanism cannot be isolated because it is intrinsic to all eight EfficientNet variants evaluated here. These questions motivate controlled future experiments rather than mechanistic conclusions from the current observations.

Overall, the findings support a conservative interpretation: within the evaluated fixed-resolution transfer-learning protocol, increasing EfficientNet scale changed computational requirements substantially but did not produce a corresponding monotonic increase in diagnostic performance. Within the evaluated protocol, B1 achieved the highest observed test accuracy and Drusen F1, whereas the larger variants exhibited partial recovery in test performance without a monotonic increase. The principal class-specific variation was concentrated in Drusen, emphasizing the value of class-level analysis when assessing architectural scaling in medical-image classification.

## 6. Conclusion

This study evaluated the complete EfficientNet-B0 → B7 family under a unified fixed-resolution transfer-learning protocol for four-class spectral-domain optical coherence tomography (SD-OCT) retinal disease classification. By holding the input resolution at 224 × 224 and applying a common preprocessing, augmentation, classification-head, and progressive fine-tuning protocol across all eight variants, the study characterized the downstream response to within-family architectural scaling under controlled experimental conditions.

The principal finding was that increasing EfficientNet scale did not produce a monotonic improvement in SD-OCT classification performance. Test accuracy increased from 95.93% for B0 to a maximum of 96.95% for B1, subsequently fluctuated across B2-B6, and recovered to 96.73% for B7. B1 therefore achieved the highest observed test accuracy, although B7 approached this performance at substantially greater model size and measured CPU inference cost. These results indicate that the increasing-scale ordering observed on ImageNet should not be assumed to transfer unchanged to the target OCT task.

Class-specific analysis further showed that the response to scaling was heterogeneous across diagnostic categories. Drusen remained the lowest-F1 class for all eight variants and accounted for 82.2% of the total summed squared deviation in class-specific F1. B1 achieved the highest Drusen F1 (0.8780), whereas B7 achieved the highest Drusen recall (0.9177) but a lower F1 (0.8720), reflecting differences in the observed precision-recall balance rather than a uniform improvement with increasing model scale.

The computational analysis showed that scaling substantially increased measured inference cost. Mean CPU latency increased from 407.48 ms for B0 to 1247.88 ms for B7, while the descriptive accuracy-to-parameter ratio decreased from 19.82 %/M to 1.48 %/M. Thus, within the evaluated configuration, the additional parameter capacity of the larger variants was not accompanied by a proportional increase in diagnostic performance.

Taken together, the findings support evaluating architectural scaling directly in the target medical-imaging domain rather than inferring downstream performance from source-domain rankings alone. Within the evaluated configuration, B1 combined the highest observed test accuracy and Drusen F1 with substantially fewer total parameters and lower measured CPU inference latency than B7. This observation should be interpreted as an empirical result within the evaluated protocol and should not be generalized to OCT classification beyond the present dataset, fixed-resolution setting, and training configuration.

The conclusions are limited using a single dataset, fixed 224 × 224 input resolution, and a single nominal random seed. Consequently, the observed scaling trajectory should be regarded as protocol and dataset-specific until it is reproduced across independent datasets, multiple random seeds, alternative resolution settings, and broader architectural families. An additional limitation is that the supplied dataset partitions were not independently audited for cross-partition duplicate images. Therefore, the independence of test-image content from training data has not been established, and the reported performance should be interpreted in this context until a file-level integrity audit is completed. Future controlled studies should also examine the interaction between target-domain data scale and model capacity to determine under what conditions larger architectures may provide a consistent downstream advantage.

**Supplementary Materials:** The following supporting information can be downloaded at https://huggingface.co/shkelqim2026/EfficientNet : Models; Figures; Tables; Scripts.

**Funding:** This research received no external funding.

**Institutional Review Board Statement:** This study used a publicly available, de-identified, pre-existing OCT dataset. The original data collection was approved by the Institutional Review Board at each contributing site (Kermany et al., 2018 [2]) and written informed consent was obtained from all subjects involved.

**Author Contributions**: Conceptualization, S.S.; methodology, S.S.; software, S.S.; validation, S.S.; formal analysis, S.S.; investigation, S.S.; data curation, S.S.; writing-original draft preparation, S.S.; writing-review and editing, S.S.; visualization, S.S. The author has read and agreed to the published version of the manuscript.

Methodology, F.I.,S.I; formal analysis, F.I.,S.I; and writing – review, F.I.,S.I.

**Informed Consent Statement:** Informed consent was obtained from all subjects involved in the original data collection. The present analysis used only de-identified retrospective data openly available in the public domain.

**Data Availability Statement:** https://huggingface.co/shkelqim2026/EfficientNet

**Conflicts of Interest:** The authors declare no conflicts of interest.

## Abbreviations

The following abbreviations are used in this manuscript:

| | |
|---|---|
| AUC | Area under the curve |
| CAD | Computer-aided diagnosis |
| CCE | Categorical cross-entropy |
| CNN | Convolutional neural network |
| CNV | Choroidal neovascularization |
| CPU | Central processing unit |
| DME | Diabetic macular edema |
| F1 | F1 score |
| GAP | Global average pooling |
| GPU | Graphics processing unit |
| Grad-CAM | Gradient-weighted Class Activation Mapping |
| MBConv | Mobile Inverted Bottleneck Convolution |
| MCC | Matthew's correlation coefficient |
| OCT | Optical coherence tomography |
| PR | Precision-recall |
| PR-AUC | Area under the precision-recall curve |
| ROC | Receiver operating characteristic |
| ROC-AUC | Area under the receiver operating characteristic curve |
| RPE | Retinal pigment epithelium |
| SD-OCT | Spectral-domain optical coherence tomography |
| SE | Squeeze-and-Excitation |
| t-SNE | t-distributed stochastic neighbor embedding |
| VRAM | Video random-access memory |

## Appendix A Supplementary Quantitative Results

### *A.1. Confusion Matrices*

This appendix provides supplementary quantitative results for the eight EfficientNet variants evaluated under the fixed-resolution transfer-learning protocol. The tables complement the principal results presented in Section 4 by providing the complete confusion matrices, parameter-count breakdowns, and class-wise ROC-AUC and PR-AUC values for EfficientNet-B0 through EfficientNet-B7.

**Table A1.** Confusion matrices for EfficientNet-B0 → B7 on the held-out test set.

| Model | True/Predicted | CNV | DME | DRUSEN | NORMAL |
|---|---|---|---|---|---|
| **B0** | **CNV** | 3567 | 34 | 140 | 5 |
| **B0** | **DME** | 15 | 1115 | 4 | 27 |
| **B0** | **DRUSEN** | 47 | 8 | 790 | 42 |
| **B0** | **NORMAL** | 3 | 54 | 66 | 5016 |
| **B1** | **CNV** | 3631 | 29 | 82 | 4 |
| **B1** | **DME** | 10 | 1117 | 5 | 29 |
| **B1** | **DRUSEN** | 32 | 10 | 806 | 39 |
| **B1** | **NORMAL** | 4 | 34 | 56 | 5045 |
| **B2** | **CNV** | 3614 | 45 | 84 | 3 |
| **B2** | **DME** | 10 | 1129 | 4 | 18 |
| **B2** | **DRUSEN** | 52 | 8 | 795 | 32 |
| **B2** | **NORMAL** | 7 | 57 | 91 | 4984 |
| **B3** | **CNV** | 3598 | 33 | 109 | 6 |
| **B3** | **DME** | 11 | 1115 | 4 | 31 |
| **B3** | **DRUSEN** | 39 | 10 | 804 | 34 |
| **B3** | **NORMAL** | 6 | 26 | 72 | 5035 |
| **B4** | **CNV** | 3568 | 34 | 136 | 8 |
| **B4** | **DME** | 7 | 1119 | 9 | 26 |
| **B4** | **DRUSEN** | 25 | 9 | 810 | 43 |
| **B4** | **NORMAL** | 4 | 55 | 80 | 5000 |
| **B5** | **CNV** | 3597 | 39 | 107 | 3 |
| **B5** | **DME** | 7 | 1128 | 4 | 22 |
| **B5** | **DRUSEN** | 38 | 9 | 802 | 38 |
| **B5** | **NORMAL** | 4 | 60 | 59 | 5016 |
| **B6** | **CNV** | 3618 | 25 | 100 | 3 |
| **B6** | **DME** | 14 | 1119 | 3 | 25 |
| **B6** | **DRUSEN** | 31 | 11 | 812 | 33 |
| **B6** | **NORMAL** | 6 | 36 | 111 | 4986 |
| **B7** | **CNV** | 3605 | 22 | 117 | 2 |
| **B7** | **DME** | 14 | 1130 | 4 | 13 |
| **B7** | **DRUSEN** | 33 | 9 | 814 | 31 |
| **B7** | **NORMAL** | 5 | 63 | 45 | 5026 |

### *A.2. Parameter-Count and Model-Size Breakdown*

**Table A2.** Parameter-count and model-size breakdown for EfficientNet-B0 → B7.

| Model | Backbone Params | Backbone Size(MB) | Custom Head Params | Custom Head Size (MB) | Total Params | Total Size (MB) |
|---|---|---|---|---|---|---|
| **B0** | 4,051,107 | 15.45 | 789,764 | 3.01 | 4,840,871 | 18.47 |
| **B1** | 6,576,775 | 25.09 | 789,764 | 3.01 | 7,366,539 | 28.10 |
| **B2** | 7,770,105 | 29.64 | 855,300 | 3.26 | 8,625,405 | 32.90 |
| **B3** | 10,785,071 | 41.14 | 920,836 | 3.51 | 11,705,907 | 44.65 |
| **B4** | 17,675,359 | 67.43 | 1,051,908 | 4.01 | 18,727,267 | 71.44 |
| **B5** | 28,515,063 | 108.78 | 1,182,980 | 4.51 | 29,698,043 | 113.29 |
| **B6** | 40,961,679 | 156.26 | 1,314,052 | 5.01 | 42,275,731 | 161.27 |
| **B7** | 64,099,223 | 244.52 | 1,445,124 | 5.51 | 65,544,347 | 250.03 |

*A.3. Per-Class ROC-AUC and PR-AUC*

**Table A3.** Per-class ROC-AUC and PR-AUC for EfficientNet-B0 → B7.

| | CNV | | DME | | DRUSEN | | NORMAL | |
|---|---|---|---|---|---|---|---|---|
| **Model** | **ROC-AUC** | **PR-AUC** | **ROC-AUC** | **PR-AUC** | **ROC-AUC** | **PR-AUC** | **ROC-AUC** | **PR-AUC** |
| **B0** | 0.998 | 0.996 | 0.995 | 0.974 | 0.985 | 0.910 | 0.997 | 0.996 |
| **B1** | 0.998 | 0.996 | 0.997 | 0.978 | 0.989 | 0.910 | 0.998 | 0.998 |
| **B2** | 0.998 | 0.996 | 0.998 | 0.980 | 0.987 | 0.885 | 0.997 | 0.997 |
| **B3** | 0.998 | 0.996 | 0.997 | 0.984 | 0.988 | 0.908 | 0.998 | 0.997 |
| **B4** | 0.998 | 0.997 | 0.998 | 0.984 | 0.987 | 0.906 | 0.998 | 0.997 |
| **B5** | 0.998 | 0.996 | 0.998 | 0.982 | 0.985 | 0.917 | 0.998 | 0.996 |
| **B6** | 0.998 | 0.996 | 0.997 | 0.984 | 0.986 | 0.902 | 0.997 | 0.996 |
| **B7** | 0.998 | 0.997 | 0.997 | 0.980 | 0.986 | 0.925 | 0.998 | 0.997 |